\documentclass[11pt,a4paper]{article}
\usepackage{times,latexsym}
\usepackage{url}
\usepackage[T1]{fontenc}
\usepackage[utf8]{inputenc}
\usepackage{CJKutf8}

\usepackage[acceptedWithA]{tacl2021v1}

\usepackage{amsmath}
\usepackage{amssymb}
\usepackage{bm}
\usepackage{enumitem}
\usepackage{booktabs}
\usepackage{graphicx}
\usepackage{xcolor}
\usepackage{pdfcolfoot}
\makeatletter\renewcommand{\ESO@HookIII}{\normalcolor}\makeatother

\font\sixly=lasy6
\def\wave{\raisebox{0.8pt}{\sixly\char58\sixly\char58\sixly\char58\sixly\char58}}

\DeclareRobustCommand{\kaomoji}[1]{#1}
\newcommand{\jpmin}[1]{\begin{CJK*}{UTF8}{min}#1\end{CJK*}}

\def\ie{{\it i.e.}}
\def\cellbox#1#2{%
\parbox[c]{#1}{\baselineskip=0.7\baselineskip\rule{0pt}{.8em}\noindent
#2\rule[-3pt]{0pt}{1em}}}

\def\mid{\kern1.1pt|\kern1.1pt}

\title{Holographic Neural PCFG for Unsupervised Parsing}

\author{
 Ryosuke Yamaki$^1$ \quad
 Daichi Mochihashi$^2$ \quad
 Nobutaka Shimada$^1$ \quad
 Tadahiro Taniguchi$^{3,1}$ \quad \\
 $^1$Ritsumeikan University, Japan \quad
 $^2$The Institute of Statistical Mathematics, Japan \\
 $^3$Kyoto University, Japan \\
 {\tt yamaki.ryosuke@em.ci.ritsumei.ac.jp},\quad {\tt daichi@ism.ac.jp} \\
 {\tt nshimada@is.ritsumei.ac.jp}, \quad {\tt taniguchi@i.kyoto-u.ac.jp}
}

\date{}

\begin{document}

\maketitle

\begin{abstract}
Unsupervised constituency parsing aims to accurately induce latent tree structures from raw text alone.
Recent neural parameterizations of PCFGs like the Neural PCFG family
achieve strong performance in both supervised and unsupervised parsing, yet score rules with high-capacity MLPs
so that the structure of grammar rules must be learned in MLP weights.
We propose Holographic Neural PCFG (Hol-PCFG), which recasts PCFG rule scoring as algebraic relation modeling among grammar-symbol embeddings.
Hol-PCFG adapts Holographic Embeddings \cite{nickel2016holographic} that scores knowledge-graph triples via circular correlation to the left-child, right-child, and lexical-emission relations over the embeddings
on a torus.
The non-commutativity of circular correlation provides an asymmetric parent--child interaction with only a $d$-dimensional relation vector per role, and the torus constraint makes the distance between symbol embeddings reflect the difference between their rule distributions, which contributes to the interpretability of the induced grammar through the embedding space.
Hol-PCFG achieves state-of-the-art parsing performance in six languages with 99.94\% fewer rule-scoring parameters than the baseline, while being trained more stably.
Hol-PCFG can also parse Japanese directly from characters without morphological segmentation, nearly matching morpheme-level performance.
\end{abstract}

\vspace{-1em}
\section{Introduction}
\label{sec:intro}
\begin{figure*}[ht]
\centering
\includegraphics[width=0.85\linewidth]{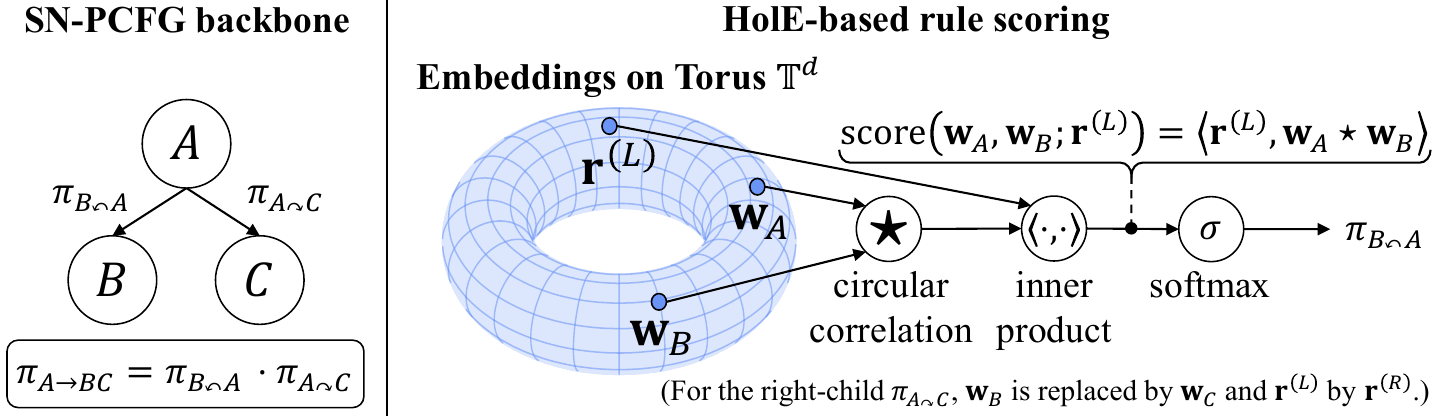}
\caption{
Overview of Hol-PCFG.
Binary rules adopt the SN-PCFG factorization, which assumes that the two children are conditionally independent given the parent $A$.
Rule probabilities are normalized HolE-based bilinear scores built on circular correlation $\star$ between torus-constrained grammar-symbol and vocabulary embeddings, with separate relation vectors $\mathbf{r}^{(L)}$, $\mathbf{r}^{(R)}$, and $\mathbf{r}^{(E)}$ for the left-child, right-child, and lexical-emission roles.
}
\label{fig:hol-pcfg-overview}
\end{figure*}

Unsupervised constituency parsing induces the syntactic tree structure of natural language from raw text
alone.
This task asks whether compact discrete grammars can be induced only from word-level distributional information. 
This question remains relevant even in the era of large language models (LLMs).
LLM-based unsupervised parsers \citep{cao-etal-2020-unsupervised,li-lu-2023-contextual,chen-etal-2024-unsupervised} can induce parse trees using rich pretrained knowledge, but they do not necessarily learn an explicit generative grammar of reusable nonterminals and rule probabilities.
PCFG-based unsupervised parsing therefore remains important, including for low-resource languages and computational models of language acquisition~\citep{bannard-etal-2009-modeling,jin-etal-2021-character-based,jin-etal-2021-depth}.

We focus on the Neural PCFG (N-PCFG;~\citealp{kim-etal-2019-compound}) family, which retains an explicit discrete grammar while parameterizing rule probabilities with neural embeddings.
However, N-PCFG gives each child pair of a binary rule its own embedding.
Thus, the parameter count 
of the rule scorer grows quadratically with the number of nonterminals and preterminals, restricting early models to a few dozen symbols.
Subsequent work showed that induction quality improves substantially with more symbols, and scaled grammars to hundreds or thousands of nonterminals.
TN-PCFG \citep{yang-etal-2021-pcfgs} did so via a low-rank decomposition of the rule tensor, and SN-PCFG \citep{liu-etal-2023-simple} via a conditional-independence assumption between the two children, together with efficient GPU computation.
In a complementary recent direction, SemInfo \citep{chen2025improving} addresses the mismatch between likelihood improvements and parsing accuracy with a new objective that directly exploits the semantic information content of induced constituents.

The structure of the rule-scoring function itself, however, has received comparatively little attention
while the advances above target grammar size and training objectives.
In all of these models, parent--child and preterminal--lexical relations are scored by high-capacity MLPs, which treat rule scoring as generic function approximation.
Yet PCFG rule scoring has an intrinsic structure, namely directionality from parent to child, asymmetry between the left and right children, and similarities among grammar symbols, \ie, which symbols have similar rule distributions.
When this structure is left implicit in MLP weights, it must be rediscovered only from the weak signal of raw-text likelihood, and this often fails.
Distinct nonterminals collapse onto nearly identical rule distributions \citep{park-kim-2025-probability}.
Moreover, the MLPs leave the norms of the transformed vectors unconstrained.
Thus, the distance between symbol embeddings is not guaranteed to reflect the difference between their rule distributions.
The central problem motivating this work is whether rule scoring in N-PCFGs can be redesigned not as generic function approximation by MLPs, but as algebraic relation scoring between grammar-symbol embeddings, carrying this structure by construction.

We propose Holographic Neural PCFG (Hol-PCFG; Figure~\ref{fig:hol-pcfg-overview}), which scores PCFG rules as algebraic relations between symbol embeddings on a high-dimensional torus.\footnote{The code is available at \url{https://github.com/ryosuke-yamaki/hol-pcfg.git}}
Hol-PCFG adapts Holographic Embeddings (HolE;~\citealp{nickel2016holographic}) that scores knowledge-graph triples via circular correlation~\citep{plate1995holographic} to the left-child, right-child, and lexical-emission relations of a PCFG.
Circular correlation is non-commutative, which encodes the direction from parent to child.
Two grammar-wide relation vectors distinguish the left and right roles.
The scorer therefore learns only one $d$-dimensional vector per role.
Moreover, the torus constraint makes the distance between symbol embeddings reflect the difference between their rule distributions, so the similarities among grammar symbols can be examined in the embedding space.
Hol-PCFG thus shows that unsupervised constituency parsing can be improved not only by larger nonterminal inventories or better training objectives, but also by a structural inductive bias in rule scoring.

The contributions of this work are threefold.
\begin{enumerate}[itemsep=2pt]
  \item We propose Hol-PCFG, which scores Neural PCFG rules with HolE-based operations over torus-constrained embeddings, so the parent--child relation is scored asymmetrically with a $d$-dimensional relation vector per role and the similarities among grammar symbols can be examined in the embedding space.
  \item Hol-PCFG achieves state-of-the-art parsing performance in six languages (English with the SemInfo objective, Chinese, French, Korean, Swedish, and Japanese), with far fewer parameters and higher training stability.
  \item We show that Hol-PCFG can even parse directly from characters without morphological segmentation while retaining nearly the same morpheme-level performance. We provide experimental results showing that it can also induce plausible hierarchical analyses of non-linguistic text-like data as well as ordinary text.
\end{enumerate}

\section{Background and Related Work}
\label{sec:background}
\subsection{Unsupervised Constituency Parsing}
\label{ssec:bg-pcfg}

Unsupervised constituency parsing methods either induce an explicit grammar with reusable nonterminal symbols and rule probabilities or only produce trees (bracketings).
Most prior methods, including recent LLM-based approaches, do the latter and differ mainly in the signal they exploit.
Classical methods induce unlabeled brackets without nonterminal categories, as in the constituent-context distributional model of \citet{klein-manning-2002-generative} and the incremental link-based parser of \citet{seginer-2007-fast}.
A second group derives trees from a neural language model via syntactic distances in PRPN \citep{shen-etal-2018-neural}, ordered neurons in ON-LSTM \citep{shen-etal-2019-ordered}, or latent trees in URNNG \citep{kim-etal-2019-unsupervised}.
DIORA and S-DIORA \citep{drozdov-etal-2019-unsupervised,drozdov-etal-2020-unsupervised} use a self-supervised inside-outside recursive autoencoder that reconstructs each word from its context.
Recent work leverages pretrained models by reading structure off their representations via learned constituency tests \citep{cao-etal-2020-unsupervised} or the contextual distortion of a frozen masked language model \citep{li-lu-2023-contextual}, or using LLM-generated paraphrases whose recurring word sequences signal constituents \citep{chen-etal-2024-unsupervised}.

In contrast, the Neural PCFG~\citep{kim-etal-2019-compound} family, including this work, induces an explicit grammar.
A probabilistic context-free grammar $\mathcal{G}$ attaches a probability $\pi$ to each production rule. Because the parse tree is unobserved, grammar induction fits $\mathcal{G}$ and $\pi$ to the data by maximizing the log marginal likelihood of a sentence $x = w_1 \cdots w_\ell$, which is computed efficiently by the inside algorithm \citep{baker-1979-trainable} and marginalizes over all parse trees $t \in T_{\mathcal{G}}(x)$ that yield $x$:
{\abovedisplayskip=0.7em \belowdisplayskip=-5pt
\begin{equation}
  \log p_{\pi}(x) = \log\!\sum_{t \in T_{\mathcal{G}}(x)}\! p_{\pi}(t,x).
  \label{eq:pcfg-ll}
\end{equation}
}

\subsection{Neural PCFG}
\label{ssec:bg-npcfg}
N-PCFG parameterizes rule probabilities $\pi$ with neural embeddings and a softmax function, using a pair embedding $u_{BC}$ for each binary rule $A \to BC$.
With $\mathcal{O}(|\mathcal{N}|(|\mathcal{N}|+|\mathcal{P}|)^2)$ binary rules for $|\mathcal{N}|$ nonterminals and $|\mathcal{P}|$ preterminals, only small grammars such as $|\mathcal{N}| \approx 30$, $|\mathcal{P}| \approx 60$ were practical.

In contrast, 
TN-PCFG \citep{yang-etal-2021-pcfgs} factorizes the rule tensor by low-rank decomposition and scales $|\mathcal{N}|$ to hundreds.
Building on this line of work, Simple Neural PCFG (SN-PCFG; \citealp{liu-etal-2023-simple}) introduces an assumption that the two children are conditionally independent given the parent $A$, and decomposes binary-rule probabilities into the product of two categorical distributions.
Each factor is parameterized by multiple MLPs that handle the parent, child, and lexical transformations of the embeddings.
SN-PCFG proposes the \mbox{FlashInside} implementation that combines kernel fusion with log-einsum-exp, and extends $|\mathcal{N}|$ to 8192.
Its variant SC-PCFG adds the per-sentence global latent vector $z$ that Compound PCFG (C-PCFG; \citealp{kim-etal-2019-compound}) adds to N-PCFG.

A further question is whether this enlarged symbol inventory is used effectively.
\citet{park-kim-2025-probability} identify \emph{probability distribution collapse}, in which distinct nonterminals have nearly identical rule distributions.
Mitigating this collapse keeps far smaller grammars competitive.
Their approach repairs the standard inner-product scorer to compress the inventory, whereas Hol-PCFG redesigns the scoring operation to compress its parameterization, taking an orthogonal route to compact~grammars.

Other extensions of N-PCFG include visual grounding \citep{zhao-titov-2020-visually}, lexicalization \citep{zhu-etal-2020-return,yang-etal-2021-neural}, and discontinuous constituency \citep{yang-etal-2023-unsupervised}.

\subsection{SemInfo Objective}
\label{ssec:bg-seminfo}
The log-likelihood of Equation~(\ref{eq:pcfg-ll}) is the standard objective for PCFG induction but has been reported to become only weakly correlated with sentence-level F1 (SF1) as training progresses.
In response, \citet{chen2025improving} proposed SemInfo training, which views a parse tree $t$ as a set of constituent substrings $t = \{s_1, s_2, \dots\}$ and maximizes the semantic information
$I(t, m(x)) = \sum_{s \in t} I(s, m(x))$
between $t$ and a semantic representation $m(x)$ of sentence $x$.
Each term extends the Probability-Weighted Information (PWI; a probabilistic interpretation of tf-idf) of \citet{aizawa2003information} to a bag of substrings: $I(s, m(x))=p(s\mid m(x))\log \bigl(p(m(x)\mid s) / p(m(x))\bigr)$.
Both factors are empirically estimated from the frequency and inverse document frequency of maximal substrings in the LLM-generated paraphrase set $X_p$.

SemInfo uses reinforcement learning with a span-based TreeCRF \citep{stern-etal-2017-minimal,kim-etal-2019-unsupervised} and is reported to significantly improve parsing performance for several N-PCFG-family models (\S\ref{ssec:bg-npcfg}) in multiple languages.
In \S\ref{sec:experiments}, we evaluate Hol-PCFG under both maximum-likelihood training and the SemInfo objective.

\subsection{Holographic Embeddings}
\label{ssec:bg-hole}
\begin{figure}[tb]
    \centering
    \includegraphics[width=\linewidth]{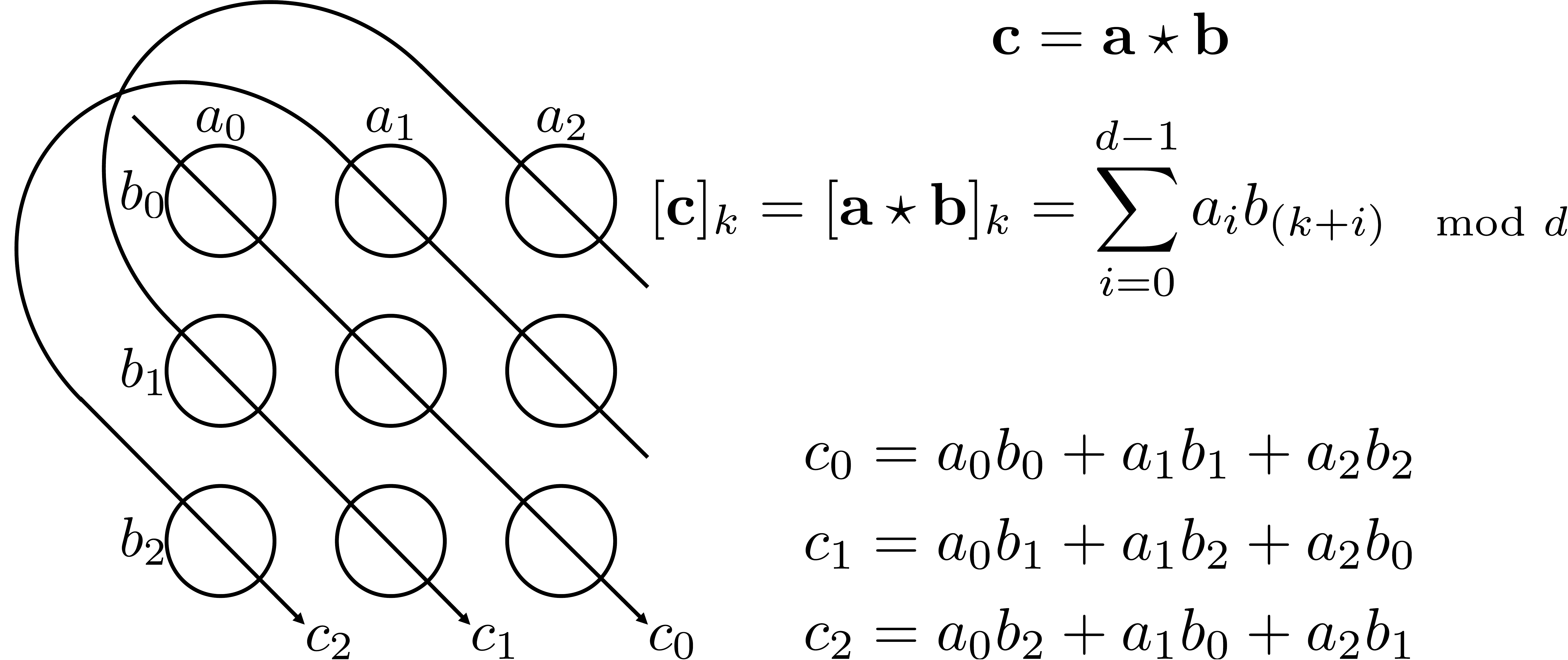}
    \caption{Circular correlation, adapted from~\citet{plate1995holographic,nickel2016holographic,yamaki-etal-2023-holographic}.
    Circles are the products $a_i b_j$.
    Arrows show the additions.\label{fig:circular_correlation}}
\end{figure}

Holographic Embeddings (HolE;~\citealp{nickel2016holographic}) is a knowledge-graph embedding method that builds on holographic reduced representations~\citep{plate1995holographic}.
Given a triple $(s,p,o)$ of a subject entity $s$, a relation $p$, and an object entity $o$,
it models the probability that $p$ holds between $s$ and $o$ as the sigmoid $\sigma$ of a bilinear score based on circular correlation~$\star$: $p\bigl({\phi_p(s,o)=1} \mid \Theta\bigr) = \sigma\!\bigl(\mathbf{r}_p^{\top}(\mathbf{e}_s \star \mathbf{e}_o)\bigr)$, where $\mathbf{e}_s, \mathbf{e}_o, \mathbf{r}_p \in \mathbb{R}^d$ are the embeddings of $s$, $o$, and $p$, and $\Theta$ is the set of all embeddings.
For two vectors $\mathbf{a}, \mathbf{b} \in \mathbb{R}^d$, the circular correlation $\star : \mathbb{R}^d \times \mathbb{R}^d \to \mathbb{R}^d$ is defined as
{\abovedisplayskip=5pt \belowdisplayskip=2pt
\begin{align}
  [\mathbf{a} \star \mathbf{b}]_k
  &= \sum_{i=0}^{d-1} a_i\, b_{(k+i) \bmod d},
  \quad k = 0, \dots, d{-}1, \notag\\[-1.2em]
  \label{eq:circular-correlation}
\end{align}
}%
and can be computed as
$\mathcal{F}^{-1}\!\bigl(\overline{\mathcal{F}(\mathbf{a})} \odot \mathcal{F}(\mathbf{b})\bigr)$ in $\mathcal{O}(d \log d)$,
where $\mathcal{F}$ is the discrete Fourier transform, $\overline{\,\cdot\,}$ is the complex conjugate, and $\odot$ is the element-wise product.
See Figure~\ref{fig:circular_correlation} for $d\!=\!3$.

Circular correlation has two properties desirable for
scoring grammatical relations among entities. (i) Non-commutativity ($\mathbf{a} \star \mathbf{b} \neq \mathbf{b} \star \mathbf{a}$) allows asymmetric roles for subject and object, and (ii) the output dimension remains $d$, so relation scoring requires only $\mathcal{O}(d)$ parameters.
With these properties, HolE achieved state-of-the-art results at the time on knowledge-graph completion with high parameter efficiency.
By Equation~\eqref{eq:circular-correlation}, the HolE score is a bilinear form, $\mathbf{r}_p^{\top}(\mathbf{e}_s \star \mathbf{e}_o) = \mathbf{e}_s^{\top} M \mathbf{e}_o$ with the circulant matrix $M_{ij} = [\mathbf{r}_p]_{(j-i) \bmod d}$.
Property (i) corresponds to $M$ not being symmetric in general, and (ii) to $M$ being determined by the $d$ entries of $\mathbf{r}_p$.
Other bilinear knowledge-graph embedding methods have only one of the two properties.
DistMult~\citep{yang2015embedding} uses $M = \mathrm{diag}(\mathbf{r}_p)$, which has $d$ entries but is symmetric, whereas RESCAL~\citep{nickel2011three} learns a general $M$ per relation, which can be asymmetric but has $d^2$ entries.
These two properties correspond closely to what PCFG rule scoring inherently requires, namely the directed parent-to-child relation in a binary rule $A \to BC$ and parameter efficiency when scaling up $|\mathcal{N}|$.
With parent and child as subject and object, and each role (left child, right child, or lexical emission) as a relation, these methods and distance-based ones such as TransE~\citep{bordes2013translating} and MuRE~\citep{balazevic2019multi} can also score PCFG rules. We compare such methods with HolE empirically in \S\ref{ssec:disc-ablation}.

\citet{yamaki-etal-2023-holographic} applied HolE to supervised CCG parsing (Hol-CCG), recursively composing phrases as $\mathbf{c} = \mathbf{a} \star \mathbf{b}$ and using a complex unit projection $\mathbf{x} \leftarrow \mathcal{F}^{-1}\bigl[(\mathcal{F}(\mathbf{x})_k / |\mathcal{F}(\mathbf{x})_k|)_{k=0}^{d-1}\bigr]$ to suppress recursive norm divergence.
Hol-CCG achieved state-of-the-art CCG supertagging and parsing accuracy at the time.
We adapt these operations to score relations among grammar-symbol embeddings for unsupervised PCFG induction.

\section{Holographic Neural PCFG}
\label{sec:method}

\begin{figure}[tb]
    \centering
    \includegraphics[width=0.8\linewidth]{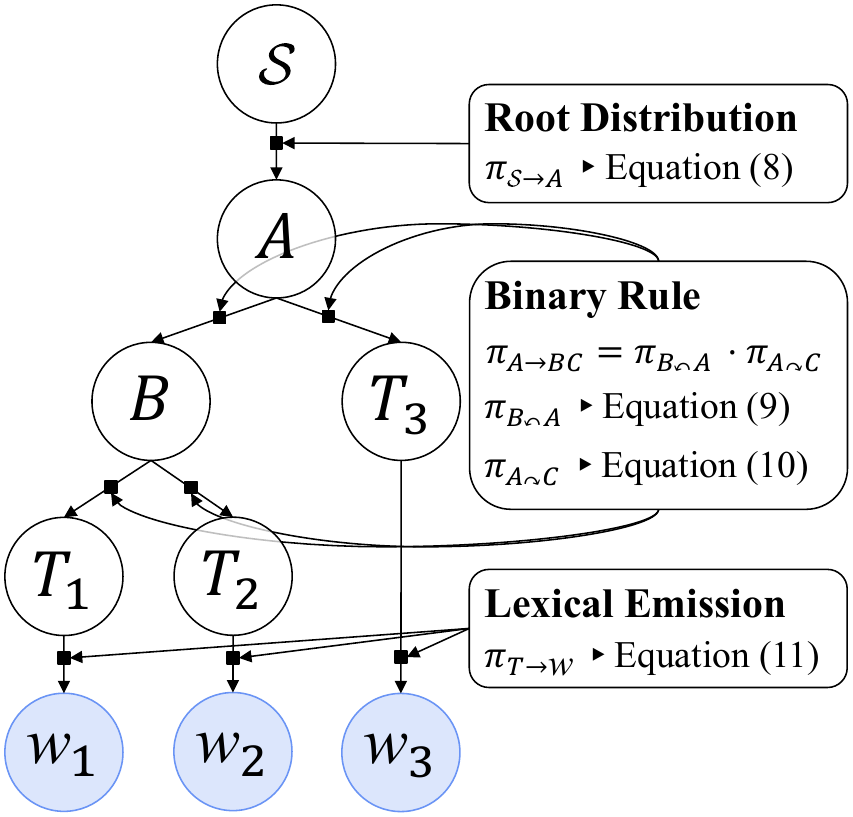}
    \caption{Schematic of Hol-PCFG for an example tree with start symbol $\mathcal{S}$, nonterminals $A,B \in \mathcal{N}$, preterminals $T_1, T_2, T_3 \in \mathcal{P}$, and terminals $w_1, w_2, w_3 \in \Sigma$.
    The root distribution uses scaled inner products.
    The binary-rule and lexical-emission distributions use HolE-based scores over torus-constrained embeddings.}
    \label{fig:method}
\end{figure}

Here, we formulate Hol-PCFG as shown in Figure~\ref{fig:method}.

\subsection{Grammar Formulation}
\label{ssec:method-grammar}
Hol-PCFG models a probabilistic context-free grammar (PCFG)
$\mathcal{G} = (\mathcal{S}, \mathcal{N}, \mathcal{P}, \Sigma, \mathcal{R}, \pi)$, where $\mathcal{S}$ is the start symbol; $\mathcal{N}$, $\mathcal{P}$, and $\Sigma$ are finite sets of nonterminal, preterminal, and terminal (vocabulary) symbols; $\mathcal{R}$ is the set of three types of production rules; and $\pi : \mathcal{R} \to [0, 1]$ maps each rule~$r$ to its probability $\pi_r$ ($\pi_{\mathcal{S} \to A}$, $\pi_{A \to BC}$, or $\pi_{T \to w}$):
\begin{align}
  \mathcal{S} &\to A,
    && A \in \mathcal{N},
    \label{eq:root_rule} \\
  A &\to B\, C,
    && A \in \mathcal{N},\; B, C \in \mathcal{N} \cup \mathcal{P},
    \label{eq:binary_rule} \\
  T &\to w,
    && T \in \mathcal{P},\; w \in \Sigma.
    \label{eq:lexical_rule}
\end{align}
A preterminal always spans exactly one terminal, whereas a nonterminal spans two or more contiguous terminals through binary branching.

Following SN-PCFG~\citep{liu-etal-2023-simple}, we assume that the choices of the left and right children are conditionally independent given the parent $A$.\footnote{Only the two siblings are independent. The parent--child interaction remains asymmetric and directed from parent to child, because circular correlation $\star$ is non-commutative (\S\ref{sssec:method-binary}).}
Let $B \curvearrowleft A$ and $A \curvearrowright C$ denote the events ``$B$ is chosen as the left child of $A$'' and ``$C$ is chosen as the right child of $A$''.
We then define
{\abovedisplayskip=5pt
\begin{equation}
  \pi_{A \to BC} = \pi_{B \curvearrowleft A} \cdot \pi_{A \curvearrowright C},
  \vspace{-0.5em}
  \label{eq:factorized_rule}
\end{equation}
}%
where $\pi_{\cdot \curvearrowleft A}$ and $\pi_{A \curvearrowright \cdot}$ are both categorical distributions over $\mathcal{N} \cup \mathcal{P}$.

\vspace{-0.5em}
\subsection{HolE-Based Rule Parameterization}
\label{ssec:method-rule-param}
Binary and lexical-emission rules use the HolE-based score, defined for $\mathbf{r}, \mathbf{a}, \mathbf{b} \in \mathbb{R}^d$ as
\vspace{-0.5em}
\begin{equation}
  \mathrm{score}(\mathbf{a}, \mathbf{b};\, \mathbf{r})
  = \bigl\langle \mathbf{r},\;
    \mathbf{a} \star \mathbf{b}
  \bigr\rangle ,
  \vspace{-0.5em}
  \label{eq:hole_score}
\end{equation}
which is high when $\mathbf{r}$ is compatible with the ordered pair $(\mathbf{a}, \mathbf{b})$ under the circular correlation $\star$ in Equation~\eqref{eq:circular-correlation}.
Hol-PCFG replaces the MLP-based factor scorer of SN-PCFG with this score.

\subsubsection{Model Parameters}
\label{sssec:method-params}
Hol-PCFG learns a start embedding $\mathbf{w}_{\mathcal{S}}$ on the torus $\mathbb{T}^d$ (\S\ref{ssec:method-torus}), symbol embeddings $\mathbf{w}_A \in \mathbb{T}^d$ for $A \in \mathcal{N} \cup \mathcal{P}$, vocabulary embeddings $\mathbf{u}_w \in \mathbb{T}^d$ for $w \in \Sigma$, relation vectors $\mathbf{r}^{(L)}, \mathbf{r}^{(R)}, \mathbf{r}^{(E)} \in \mathbb{T}^d$ for left-child, right-child, and lexical-emission relations, and scaling parameters $\alpha, \tau, \eta \in \mathbb{R}^+$ for the root, binary-rule, and lexical-emission scores.

\subsubsection{Root Distribution}
\label{sssec:method-root}
The root distribution is computed by softmax over scaled inner products:
\vspace{-1em}
\begin{equation}
  \pi_{\mathcal{S} \to A}
  = \frac{
    \exp\bigl(
      \alpha \cdot
      \mathbf{w}_A^{\top} \mathbf{w}_{\mathcal{S}}
    \bigr)
  }{
    \displaystyle\sum_{A' \in \mathcal{N}}
    \exp\bigl(
      \alpha \cdot
      \mathbf{w}_{A'}^{\top} \mathbf{w}_{\mathcal{S}}
    \bigr)
  }.
  \label{eq:root_prob}
\end{equation}

\subsubsection{Binary Rule Probabilities}
\label{sssec:method-binary}
Left- and right-child distributions are computed from $\mathbf{r}^{(L)}$ and $\mathbf{r}^{(R)}$:
{\abovedisplayskip=5pt
\begin{align}
  \pi_{B \curvearrowleft A}
  &= \frac{
    \exp\bigl(
      \tau \cdot
      \langle \mathbf{r}^{(L)},\,
        \mathbf{w}_A \star \mathbf{w}_B
      \rangle
    \bigr)
  }{
    \displaystyle\sum_{B' \in \mathcal{N} \cup \mathcal{P}}
    \exp\bigl(
      \tau \cdot
      \langle \mathbf{r}^{(L)},\,
        \mathbf{w}_A \star \mathbf{w}_{B'}
      \rangle
    \bigr)
  }, \notag\\[-1em]
  \label{eq:left_child_prob} \\
  \pi_{A \curvearrowright C}
  &= \frac{
    \exp\bigl(
      \tau \cdot
      \langle \mathbf{r}^{(R)},\,
        \mathbf{w}_A \star \mathbf{w}_C
      \rangle
    \bigr)
  }{
    \displaystyle\sum_{C' \in \mathcal{N} \cup \mathcal{P}}
    \exp\bigl(
      \tau \cdot
      \langle \mathbf{r}^{(R)},\,
        \mathbf{w}_A \star \mathbf{w}_{C'}
      \rangle
    \bigr)
  }. \notag\\[-1em]
  \label{eq:right_child_prob}
\end{align}
}

In both factors, the parent embedding $\mathbf{w}_A$ is the first argument of $\star$, so the non-commutativity of circular correlation encodes the directed parent-to-child relation rather than the distinction between left and right siblings, separating $A$ as a parent from $A$ as a child.
The left and right roles are carried by $\mathbf{r}^{(L)}$ and $\mathbf{r}^{(R)}$, which score the parent--child pairs $(A,B)$ and $(A,C)$, respectively.
The non-associativity of circular correlation ($(\mathbf{a} \star \mathbf{b}) \star \mathbf{c} \neq \mathbf{a} \star (\mathbf{b} \star \mathbf{c})$) is not essential to Hol-PCFG, because the scores in Equations~\eqref{eq:left_child_prob} and~\eqref{eq:right_child_prob} each compose only two vectors, $\mathbf{w}_A$ and one child embedding.

\subsubsection{Lexical Emission Probabilities}
\label{sssec:method-lex}
Lexical emissions use the relation vector $\mathbf{r}^{(E)}$:
\begin{equation}
  \pi_{T \to w}
  = \frac{
    \exp\bigl(
      \eta \cdot
      \langle \mathbf{r}^{(E)},\,
        \mathbf{w}_T \star \mathbf{u}_w
      \rangle
    \bigr)
  }{
    \displaystyle\sum_{w' \in \Sigma}
    \exp\bigl(
      \eta \cdot
      \langle \mathbf{r}^{(E)},\,
        \mathbf{w}_T \star \mathbf{u}_{w'}
      \rangle
    \bigr)
  }.
  \label{eq:emission_prob}
\end{equation}

\subsubsection{Efficient Scoring}
\label{sssec:method-efficient}
To evaluate the scores in Equations~\eqref{eq:left_child_prob}--\eqref{eq:emission_prob} efficiently, we use a  circular convolution $*$:\footnote{Circular correlation $\star$ and circular convolution $*$ are related but distinct. We distinguish them throughout this paper.}
{\abovedisplayskip=5pt \belowdisplayskip=5pt
\begin{equation}
  (\mathbf{r} * \mathbf{a})_n
  = \!\sum_{m=0}^{d-1} r_m \, a_{(n-m)\bmod d}.
  \label{eq:circ_conv_def}
\end{equation}
}%
Then, the HolE score can be rewritten as
{\abovedisplayskip=5pt \belowdisplayskip=5pt
\begin{equation}
  \langle \mathbf{r},\, \mathbf{a} \star \mathbf{b} \rangle
  = \langle \mathbf{b},\, \mathbf{r} * \mathbf{a} \rangle ,
  \label{eq:corr_conv_identity}
\end{equation}
}%
which follows from Equations~\eqref{eq:circular-correlation} and~\eqref{eq:circ_conv_def} by changing the summation index (Appendix~\ref{app:efficient-scoring}).
The convolution is computed as $\mathbf{r} * \mathbf{a} = \mathcal{F}^{-1}\!\bigl(\mathcal{F}(\mathbf{r}) \odot \mathcal{F}(\mathbf{a})\bigr)$ in $\mathcal{O}(d \log d)$.
For the left-child scores in Equation~\eqref{eq:left_child_prob}, the vector $\mathbf{r}^{(L)} * \mathbf{w}_A$ on the right-hand side of Equation~\eqref{eq:corr_conv_identity} depends only on $A$, so it is computed once per parent and shared by all candidate children.
We call it the \emph{template} of $A$ (the right-child and lexical-emission scores use $\mathbf{r}^{(R)} * \mathbf{w}_A$ and $\mathbf{r}^{(E)} * \mathbf{w}_T$ likewise).
Compared with computing the left-hand side of Equation~\eqref{eq:corr_conv_identity} for every parent--child pair, this reduces both the number of inverse FFTs and the size of the materialized intermediate tensor by a factor of $|\mathcal{N}|+|\mathcal{P}|$ (Appendix~\ref{app:efficient-scoring}).

\subsection{Embeddings on a Torus}
\label{ssec:method-torus}
All embeddings in Hol-PCFG are constrained to a torus $\mathbb{T}^d \subset \mathbb{R}^d$, the set of vectors whose frequency components have unit amplitude:
\begin{equation}
  \mathbb{T}^d = \bigl\{
    \mathbf{x} \in \mathbb{R}^d \;\big|\;
    |\hat{x}_k| = 1,\;\forall\, k = 0, 1, \dots, d{-}1
  \bigr\}.
  \label{eq:torus}
\end{equation}
Here, $\hat{\mathbf{x}} = \mathcal{F}[\mathbf{x}] \in \mathbb{C}^d$ is the discrete Fourier transform (DFT) of $\mathbf{x}$.
Appendix~\ref{app:torus} summarizes the DFT facts we use (see also \citet{hayashi-shimbo-2017-equivalence} for frequency-domain HolE).
The DFT of a real vector satisfies Hermitian symmetry, ${\hat{x}_{d-k} = \overline{\hat{x}_k}}$.
With $|\hat{x}_k| = 1$, the phases $\phi_k$ of $\hat{x}_k = e^{i\phi_k}$ thus determine $\mathbf{x} \in \mathbb{T}^d$ up to the signs of at most two real-valued components.%
\footnote{The components $\hat{x}_0$ and, for even $d$, $\hat{x}_{d/2}$ are real for every real vector, so on $\mathbb{T}^d$ each is $+1$ or $-1$. See Appendix~\ref{app:torus} for the precise structure of $\mathbb{T}^d$ and why we call it a torus.}
Each embedding has roughly $d/2$ degrees of freedom.
This halving is used only for checkpoint storage and does not change the number of learned parameters (\S\ref{ssec:disc-params}).

Every vector on $\mathbb{T}^d$ has unit $\ell_2$ norm, and $\mathbb{T}^d$ is closed under circular correlation and convolution, \ie, $\mathbf{a} \star \mathbf{b},\, \mathbf{a} * \mathbf{b} \in \mathbb{T}^d$ for all $\mathbf{a}, \mathbf{b} \in \mathbb{T}^d$ (Appendix~\ref{app:torus}).

\paragraph{Advantages of the Torus Constraint.}
By Equation~\eqref{eq:corr_conv_identity}, the logit in Equation~\eqref{eq:left_child_prob} is $\tau$ times the inner product of the child embedding $\mathbf{w}_B$ and the template $\mathbf{r}^{(L)} * \mathbf{w}_A$ of the parent $A$, with the angle $\theta_{AB}$ between them:
{\abovedisplayskip=7pt \belowdisplayskip=7pt
\begin{align}
  &\tau\, \langle \mathbf{w}_B,\, \mathbf{r}^{(L)} * \mathbf{w}_A \rangle \notag \\
  &\quad = \tau\, \lVert \mathbf{r}^{(L)} * \mathbf{w}_A \rVert\, \lVert \mathbf{w}_B \rVert \cos\theta_{AB}.
  \label{eq:logit-decomposition}
\end{align}
}%
On $\mathbb{T}^d$, both vectors have unit norm, so the logit is $\tau \cos\theta_{AB}$.
The embeddings determine only which children a parent prefers (the peak position of the distribution), and $\tau$, shared by all parents, alone determines how sharply the distribution peaks (the sharpness).
Since the same holds for the right-child, lexical-emission, and root distributions, sharp distributions require the scaling parameters.
MLP-based scorers such as that of SN-PCFG leave the vector norms unconstrained, so the logit scale varies across parents and children, and this separation is not guaranteed.
Moreover, convolution with $\mathbf{r}^{(L)}$ or $\mathbf{r}^{(R)}$ shifts the phases of every embedding by those of the relation vector, a translation on $\mathbb{T}^d$ that preserves distances.%
\footnote{$\mathbb{T}^d$ lies on the unit sphere, on which the same convolution rotates the $k$-th frequency component by the phase of $\hat{r}_k$. Points on the sphere can distribute their amplitude unevenly over the frequencies, so this rotation moves different points by different distances. On $\mathbb{T}^d$, where every amplitude is 1, it moves every embedding by the same distance.}
The distance between the embeddings of two symbols thus equals the distance between their templates and reflects the difference in the peak positions of their rule distributions, not in their sharpness.
Symbols that prefer the same children can therefore be found from the positions of their embeddings, which contributes to the interpretability of the induced grammar through the embedding space (Appendix~\ref{app:torus}).

\paragraph{Initialization.}
Embeddings are initialized on $\mathbb{T}^d$ by sampling phases independently from a uniform distribution.
For the frequency components that are not real-valued,
$\phi_k \sim \mathrm{Uniform}(-\pi, \pi)$, $\hat{x}_k = e^{i\phi_k}$, and $\hat{x}_{d-k} = \overline{\hat{x}_k}$.
The real-valued components are initialized with random signs.
The real-space embedding is then $\mathbf{x} = \mathcal{F}^{-1}[\hat{\mathbf{x}}]$.

\paragraph{Projection.}
Following the complex unit projection introduced by Hol-CCG \citep{yamaki-etal-2023-holographic}, after each training update we project every embedding back onto $\mathbb{T}^d$ by renormalizing each frequency component to unit amplitude (${\hat{x}_k \leftarrow \hat{x}_k / |\hat{x}_k|}$, ${\mathbf{x} \leftarrow \mathcal{F}^{-1}[\hat{\mathbf{x}}]}$).

\subsection{Training}
\label{ssec:method-training}
Hol-PCFG minimizes the negative log-\hspace{0pt}likelihood of a corpus via backpropagation:
{\abovedisplayskip=5pt \belowdisplayskip=5pt
\begin{equation}
  \mathcal{L}
  = -\frac{1}{M}\sum_{m=1}^{M} \log p_{\pi}\bigl(x^{(m)}\bigr).
  \label{eq:nll_loss}
\end{equation}
}%
Because Hol-PCFG retains the SN-PCFG factorization, the inside computation decouples the sum over child pairs and directly reuses the efficient FlashInside GPU implementation of SN-PCFG, keeping marginalization tractable for large $|\mathcal{N}|$.
We also train and evaluate Hol-PCFG with SemInfo (\S\ref{ssec:bg-seminfo}).

\section{Experiments}
\label{sec:experiments}

\begin{table}[tb]
\centering
{\scriptsize
\begin{tabular}{ll}
\toprule
Method & SF1 \\
\midrule
Left Branching (LB)
  & $8.7$ \\
Right Branching (RB)
  & $39.5$ \\
{\it Upper Bound} (UB)
  & \textit{84.3} \\
\midrule
\multicolumn{2}{l}{\textit{Non-explicit grammar methods}} \\
PRPN~\citep{shen-etal-2018-neural}
  & $37.4$ \\
ON-LSTM~\citep{shen-etal-2019-ordered}
  & $47.7$ \\
URNNG~\citep{kim-etal-2019-unsupervised}
  & $40.7$ \\
S-DIORA~\citep{drozdov-etal-2020-unsupervised}
  & $57.6$ \\
Constituency Test~\citep{cao-etal-2020-unsupervised}
  & $62.8$ \\
StructFormer~\citep{shen-etal-2021-structformer}
  & $54.0$ \\
Fast-R2D2~\citep{hu-etal-2022-fast}
  & $57.2$ \\
SpanOverlap~\citep{chen-etal-2024-unsupervised}
  & $52.9$ \\
\midrule
\multicolumn{2}{l}{\textit{Neural PCFG family (maximum-likelihood training)}} \\
N-PCFG~\citep{kim-etal-2019-compound}
  & $50.8$ \\
C-PCFG~\citep{kim-etal-2019-compound}
  & $55.2$ \\
NL-PCFG~\citep{zhu-etal-2020-return}
  & $57.3$ \\
TN-PCFG~\citep{yang-etal-2021-pcfgs}
  & $57.7$ \\
NBL-PCFG~\citep{yang-etal-2021-neural}
  & $60.4$ \\
Rank-PCFG~\citep{yang2022dynamic}
  & $\underline{64.1}$ \\
SN-PCFG~\citep{liu-etal-2023-simple}
  & $59.7_{\pm 4.3}$ \\
SC-PCFG~\citep{liu-etal-2023-simple}
  & $58.3_{\pm 4.4}$ \\
{\bf Hol-PCFG} (this work)
  & $\mathbf{64.6}_{\pm 0.4}$ \\
\midrule
\multicolumn{2}{l}{\textit{Neural PCFG family (SemInfo training)}} \\
N-PCFG$^\dagger$~\citep{chen2025improving}
  & $63.6_{\pm 1.1}$ \\
SN-PCFG$^\dagger$~\citep{chen2025improving}
  & $\underline{67.3}_{\pm 0.5}$ \\
SC-PCFG$^\dagger$~\citep{chen2025improving}
  & $66.2_{\pm 0.5}$ \\
{\bf Hol-PCFG$^\dagger$} (this work)
  & $\mathbf{68.1}_{\pm 0.5}$ \\

\bottomrule
\end{tabular}
}
\vspace{0.5em}
\caption{PTB English test SF1. Hol-PCFG results use five seeds, and $\dagger$ marks SemInfo training.
Non-$\dagger$ models use maximum likelihood unless their original papers use a different objective.
We replicated the closest baselines SN-PCFG (whose rule factorization Hol-PCFG inherits)
and its compound variant SC-PCFG with their public implementations
and hyperparameters over five seeds (original SN/SC-PCFG scores: 65.1/60.6 under maximum-likelihood and 66.8/66.9 under SemInfo training).
All other baseline scores are from the cited papers.
Within each Neural PCFG block, the best score is in \textbf{bold} and the second-best \underline{underlined}.}
\label{tab:results}
\end{table}

\begin{table*}[tb]
\centering
\begingroup
\scriptsize
\setlength{\tabcolsep}{2pt}
\resizebox{\textwidth}{!}{%
\begin{tabular}{lllllllllll|cc}
\toprule
Method & Chinese & French & German & Basque & Hebrew & Hungarian & Korean & Polish & Swedish & Japanese & Mean & Avg. rank \\
\midrule
LB
  & $\phantom{0}9.7$
  & $\phantom{0}5.7$
  & $10.0$
  & $17.9$
  & $\phantom{0}8.5$
  & $13.3$
  & $18.5$
  & $10.9$
  & $\phantom{0}8.4$
  & $29.4$ & $13.2$ & \multicolumn{1}{c}{--} \\
RB
  & $20.0$
  & $26.4$
  & $14.7$
  & $15.4$
  & $30.0$
  & $12.7$
  & $19.2$
  & $34.2$
  & $30.4$
  & $\phantom{0}9.8$ & $21.3$ & \multicolumn{1}{c}{--} \\
UB
  & \textit{81.1}
  & \textit{72.1}
  & \textit{56.7}
  & \textit{63.2}
  & \textit{81.1}
  & \textit{58.7}
  & \textit{91.1}
  & \textit{63.8}
  & \textit{64.4}
  & \textit{76.5} & \textit{70.9} & \multicolumn{1}{c}{--} \\
\midrule
N-PCFG
  & ${26.3_{\pm 2.5}}^{\mathrm{Y}}$
  & ${45.0_{\pm 2.0}}^{\mathrm{Y}}$
  & ${42.3_{\pm 1.6}}^{\mathrm{Y}}$
  & ${35.1_{\pm 2.0}}^{\mathrm{Y}}$
  & ${45.7_{\pm 2.2}}^{\mathrm{Y}}$
  & ${43.5_{\pm 1.2}}^{\mathrm{Y}}$
  & ${28.4_{\pm 6.5}}^{\mathrm{Y}}$
  & ${43.2_{\pm 0.8}}^{\mathrm{Y}}$
  & ${17.0_{\pm 9.9}}^{\mathrm{Y}}$
  & ${29.2_{\pm 21.9}}$ & $35.6$ & $5.2$ \\
C-PCFG
  & ${38.7_{\pm 6.6}}^{\mathrm{Y}}$
  & ${45.0_{\pm 1.1}}^{\mathrm{Y}}$
  & ${43.5_{\pm 1.2}}^{\mathrm{Y}}$
  & ${36.0_{\pm 1.2}}^{\mathrm{Y}}$
  & ${45.2_{\pm 0.5}}^{\mathrm{Y}}$
  & ${\mathbf{44.9}_{\pm 1.5}}^{\mathrm{Y}}$
  & ${30.5_{\pm 4.2}}^{\mathrm{Y}}$
  & ${43.8_{\pm 1.3}}^{\mathrm{Y}}$
  & ${33.0_{\pm 15.4}}^{\mathrm{Y}}$
  & ${29.9_{\pm 23.1}}$ & $39.1$ & $4.4$ \\
TN-PCFG
  & ${39.2_{\pm 5.0}}^{\mathrm{Y}}$
  & ${39.1_{\pm 4.1}}^{\mathrm{Y}}$
  & ${47.1_{\pm 1.7}}^{\mathrm{Y}}$
  & ${36.0_{\pm 3.0}}^{\mathrm{Y}}$
  & ${39.2_{\pm 10.7}}^{\mathrm{Y}}$
  & ${43.1_{\pm 1.1}}^{\mathrm{Y}}$
  & ${\underline{35.4}_{\pm 2.8}}^{\mathrm{Y}}$
  & ${\mathbf{48.6}_{\pm 3.1}}^{\mathrm{Y}}$
  & ${40.0_{\pm 4.8}}^{\mathrm{Y}}$
  & ${56.6_{\pm 1.7}}$ & $42.4$ & $3.8$ \\
SN-PCFG
  & $\underline{40.6}_{\pm 5.2}$
  & $\underline{50.4}_{\pm 2.6}$
  & $\mathbf{49.4}_{\pm 0.6}$
  & $\mathbf{37.2}_{\pm 3.9}$
  & $\underline{48.9}_{\pm 3.9}$
  & $42.5_{\pm 1.0}$
  & $31.8_{\pm 4.3}$
  & $44.5_{\pm 3.2}$
  & $\underline{42.9}_{\pm 4.7}$
  & $\underline{58.1}_{\pm 1.6}$ & $\underline{44.6}$ & $\underline{2.5}$ \\
SC-PCFG
  & $40.4_{\pm 8.1}$
  & $48.8_{\pm 1.9}$
  & $47.1_{\pm 1.8}$
  & $\underline{36.4}_{\pm 4.2}$
  & $\mathbf{50.0}_{\pm 2.8}$
  & $35.1_{\pm 11.6}$
  & $34.6_{\pm 5.9}$
  & $46.6_{\pm 2.0}$
  & $42.0_{\pm 5.5}$
  & $56.9_{\pm 4.6}$ & $43.8$ & $3.1$ \\
{\bf Hol-PCFG}
  & $\mathbf{51.8}_{\pm 1.6}$
  & $\mathbf{50.9}_{\pm 1.6}$
  & $\underline{49.0}_{\pm 0.3}$
  & $33.7_{\pm 1.6}$
  & $45.6_{\pm 2.6}$
  & $\underline{43.9}_{\pm 0.8}$
  & $\mathbf{37.3}_{\pm 1.2}$
  & $\underline{47.2}_{\pm 0.6}$
  & $\mathbf{44.6}_{\pm 1.3}$
  & $\mathbf{59.5}_{\pm 0.3}$ & $\mathbf{46.4}$ & $\mathbf{2.1}$ \\
\bottomrule
\end{tabular}%
}
\endgroup
\caption{SF1 on multilingual datasets.
Superscript $\mathrm{Y}$ marks scores from \citet{yang-etal-2021-pcfgs}.
Unmarked values are from our five-seed runs under the same preprocessing and evaluation protocol.
As in Table~\ref{tab:results}, we replicated the closest baselines SN-PCFG and its compound variant SC-PCFG on all ten languages
(\citet{liu-etal-2023-simple} originally reported SN/SC-PCFG scores of 39.9/42.9 on Chinese, 38.0/49.9 on French, and 46.7/49.1 on German).
In each column, the best score among the PCFG models is in \textbf{bold} and the second-best \underline{underlined}.}
\label{tab:results-multilingual}
\end{table*}
\subsection{Setup}
\label{ssec:exp-setup}
\paragraph{Datasets and Splits.}
We use the Penn Treebank (PTB; English,~\citealp{marcus-etal-1993-building}) as a standard benchmark.
For multilingual evaluation, we use the Chinese Treebank (CTB;~\citealp{xue-etal-2005-penn}), Statistical Parsing of Morphologically Rich Languages (SPMRL) shared-task datasets (French, German, Basque, Hebrew, Hungarian, Korean, Polish, and Swedish;~\citealp{seddah-etal-2013-overview}), and the Keyaki Treebank (KTB; Japanese,~\citealp{butler2012keyaki}).
On PTB, we evaluate Hol-PCFG with and without the SemInfo objective using the paraphrase set released by \citet{chen2025improving}.
We use the standard train/dev/test splits of each corpus, and preprocessing such as POS tag and punctuation removal follows prior N-PCFG-based studies.

\paragraph{Training Configurations.}
Following the corresponding N-PCFG-family baselines, we set $|\mathcal{N}| = 4096$ for the maximum-likelihood setting and $|\mathcal{N}| = 1024$ for the SemInfo setting, and $d = 512$ for both. As in prior work, $|\mathcal{P}| = 2|\mathcal{N}|$.
We tune hyperparameters with Optuna~\citep{akiba2019optuna} on PTB separately for each setting and apply the resulting configurations to all languages.

\subsection{Parsing Performance}
\label{ssec:exp-parsing}
Tables~\ref{tab:results} and~\ref{tab:results-multilingual} report sentence-level unlabeled F1 (SF1) on the PTB and multilingual test splits, respectively.
Under maximum-likelihood training, Hol-PCFG achieves the highest SF1 among the Neural PCFG models, outperforming our replication of SN-PCFG.
With the SemInfo objective, Hol-PCFG further achieves state-of-the-art performance.
In the multilingual experiments, Hol-PCFG achieves state-of-the-art performance among existing N-PCFG-based models on five of ten languages, namely Chinese, French, Korean, Swedish, and Japanese.
It also obtains the highest mean SF1 and best average rank among the compared N-PCFG-based models, showing consistently strong performance across diverse languages.
The standard deviation of Hol-PCFG is also smaller than that of existing methods on many languages, indicating high stability in training.
This stability may stem from the torus constraint on all Hol-PCFG embeddings (\S\ref{ssec:disc-ablation}).

Analyzing the four main languages by gold constituent label and length (Appendix~\ref{app:lang-breakdown}), we found that Hol-PCFG is better than SN-PCFG at recovering PPs, clauses, and long constituents in English, and constituents of almost all labels and lengths in Chinese.
In French, SN-PCFG is better on NPs and slightly better on constituents of 2--4 words, whereas Hol-PCFG is better on coordinations, relative and subordinate clauses, and constituents of seven or more words.
In German, SN-PCFG is better on VPs, \emph{zu}-infinitives, and constituents of four or more words, whereas Hol-PCFG is better on adjective and adverbial phrases.
In both languages, these differences largely offset each other, leaving SF1 differences of less than one point.
Further comparisons of test log-likelihood, training data size, and model capacity (Appendix~\ref{app:ll-data-capacity}) suggest that these differences in SF1 stem from optimization, which led the two models to different grammars that fit the test data equally well.

\section{Discussion}
\label{sec:discussion}
\begin{table}[t]
\centering
\scriptsize
\setlength{\tabcolsep}{4pt}
\begin{tabular}{@{}lll@{}}
\toprule
Variant & LL & SemInfo \\
\midrule
Full Hol-PCFG                                                                  & $\mathbf{64.6}_{\pm 0.4}$ & $\mathbf{68.1}_{\pm 0.5}$ \\
\midrule
\multicolumn{3}{@{}l}{Scoring function} \\
\quad Hadamard product ($\langle \mathbf{r},\, \mathbf{a} \odot \mathbf{b}\rangle$)   & $50.9_{\pm 1.6}$ & $62.8_{\pm 0.3}$ \\
\quad Circular convolution ($\langle \mathbf{r},\, \mathbf{a} * \mathbf{b}\rangle$)    & $52.3_{\pm 2.2}$ & $63.0_{\pm 0.5}$ \\
\quad Shuffled circular convolution & $60.6_{\pm 6.1}$ & $66.9_{\pm 0.1}$ \\
\quad\quad ($\langle \mathbf{r},\, p_1(\mathbf{a}) * p_2(\mathbf{b})\rangle$) & & \\
\midrule
\multicolumn{3}{@{}l}{Knowledge-graph embedding scorer} \\
\quad RESCAL ($\mathbf{a}^{\top} R\, \mathbf{b}$) & $61.6_{\pm 5.2}$ & $67.3_{\pm 0.7}$ \\
\quad ComplEx ($\mathrm{Re}\,\langle \mathbf{r},\, \mathbf{a},\, \overline{\mathbf{b}} \rangle$) & $45.4_{\pm 2.8}$ & $65.0_{\pm 1.2}$ \\
\quad TransE ($-\lVert \mathbf{a} + \mathbf{r} - \mathbf{b} \rVert$) & $39.9_{\pm 1.3}$ & $57.2_{\pm 0.8}$ \\
\quad MuRE ($-\lVert \boldsymbol{\rho} \odot \mathbf{a} + \mathbf{r} - \mathbf{b} \rVert^2 + \beta_{\mathbf{b}}$) & $45.9_{\pm 3.7}$ & $61.6_{\pm 1.0}$ \\
\midrule
\multicolumn{1}{@{}l}{w/o scaling parameters}                               & $23.2_{\pm 1.7}$ & $34.1_{\pm 0.1}$ \\
\bottomrule
\end{tabular}
\caption{Ablation of the scoring function and scaling parameters on PTB (test SF1 over five random seeds). LL: maximum-likelihood training.}
\label{tab:ablation}
\end{table}

\begin{table}[t]
\centering
\scriptsize
\setlength{\tabcolsep}{4pt}
\begin{tabular}{@{}llll@{}}
\toprule
Initialization & Projection & LL & SemInfo \\
\midrule
Torus  & Torus (Full Hol-PCFG)        & $\mathbf{64.6}_{\pm 0.4}$ & $\mathbf{68.1}_{\pm 0.5}$ \\
       & Soft-torus penalty           & $56.3_{\pm 8.1}$ & $66.3_{\pm 0.2}$ \\
       & None                         & $49.8_{\pm 3.9}$ & $66.3_{\pm 1.7}$ \\
\midrule
Xavier & Unit $\ell_2$ sphere         & $63.4_{\pm 1.6}$ & $66.1_{\pm 0.7}$ \\
       & None                         & $58.8_{\pm 2.6}$ & $66.4_{\pm 1.1}$ \\
\bottomrule
\end{tabular}
\caption{Ablation of embedding initialization and projection on PTB (test SF1 over five random seeds).}
\label{tab:ablation-geometry}
\end{table}

\subsection{Ablation Study}
\label{ssec:disc-ablation}

\paragraph{HolE vs.\ Alternative Algebraic Scoring Functions.}
We replace the HolE-based score in Equation~\eqref{eq:hole_score} with the Hadamard product of
$\langle \mathbf{r},\, \mathbf{a} \odot \mathbf{b} \rangle$, 
circular convolution of ${\langle \mathbf{r},\, \mathbf{a} * \mathbf{b} \rangle}$, 
and shuffled circular convolution of $\langle \mathbf{r},\, p_1(\mathbf{a}) * p_2(\mathbf{b}) \rangle$ \citep{zanzotto2012distributed}, where $p_1$ and $p_2$ are distinct fixed permutations.
All three variants use exactly as many parameters as the full model (one relation vector per role).
Thus, the comparison isolates the effect of the algebraic structure of the scoring operation from that of model capacity.
The Hadamard product and circular convolution, both commutative, underperform the full model by 12--14 SF1 points under maximum-likelihood training and roughly 5 under \mbox{SemInfo}, whereas shuffled circular convolution falls short by 4.0 and 1.2 points, respectively (Table~\ref{tab:ablation}).
A likely factor behind these differences is commutativity.
Commutative operations are less suited to encoding the asymmetric structure in which a parent generates its children.
In contrast, the permutations make shuffled circular convolution non-commutative, like the circular correlation in HolE ($\mathbf{a} \star \mathbf{b} \neq \mathbf{b} \star \mathbf{a}$), and it degrades much less than the two commutative operations.
Even so, it falls short of HolE under both objectives.%
\vspace{-0.5em}
\paragraph{HolE vs.\ Knowledge-Graph Embedding Scorers.}
We also replace the HolE-based scores of the binary and lexical-emission rules with those of four knowledge-graph embedding methods competing with HolE (Table~\ref{tab:ablation}; details in Appendix~\ref{app:kg-scorers}), namely RESCAL~\citep{nickel2011three}, \mbox{ComplEx}~\citep{trouillon2016complex}, TransE~\citep{bordes2013translating}, and MuRE~\citep{balazevic2019multi}.
HolE-based scoring outperforms all four methods under both objectives.
Scorers with an asymmetric parent--child interaction (HolE, RESCAL, and ComplEx) tend to outperform those with a symmetric interaction (TransE and MuRE), consistent with the discussion of commutativity above.
Shuffled circular convolution and RESCAL also have this asymmetry and perform close to HolE, but HolE achieves the highest SF1 of the three, with only $d$ parameters per role ($d^2$ for RESCAL).%
\vspace{-0.5em}
\paragraph{Learnable Scaling Parameters.}
We found that fixing all scaling parameters to~1 severely degrades SF1, by about 41 points under maximum-likelihood training and about 34 under SemInfo (Table~\ref{tab:ablation}).
The unit-norm embeddings on the torus cannot change the magnitude of the logits (\S\ref{ssec:method-torus}).
Thus, only the scaling parameters determine the sharpness of the rule distributions.
With the scaling parameters fixed to~1, every logit lies in $[-1, 1]$ and no rule distribution can be sharper.
All other ablations keep the scaling parameters, which is why this ablation has the largest effect.
Adding the same learnable scaling parameters to the scoring of SN-PCFG and SC-PCFG does not improve SF1 (Appendix~\ref{app:baseline-scaling}).
Unlike in Hol-PCFG, the MLP-based scores of SN-PCFG and SC-PCFG are unbounded, so the embeddings and MLP weights alone can adjust the score scale, making the scaling parameters redundant.
The advantage of Hol-PCFG over SN-PCFG and SC-PCFG thus stems not from the scaling parameters alone but from HolE-based scoring of torus embeddings.%
\vspace{-0.5em}
\paragraph{Initialization and Projection of Embeddings.}
In Table~\ref{tab:ablation-geometry}, we vary the embedding initialization and the projection applied after every parameter update, which maps each embedding back onto the torus (\S\ref{ssec:method-torus})
in the full Hol-PCFG.
With torus initialization, we compare the torus projection, a soft-torus penalty that replaces the projection with a loss on the squared distance of each embedding from the torus, and no projection.
With Xavier uniform initialization~\citep{glorot2010understanding}, a common choice for N-PCFG-family models~\citep{kim-etal-2019-compound,yang-etal-2021-pcfgs,liu-etal-2023-simple}, we compare projection onto the unit $\ell_2$ sphere and no projection.
Under torus initialization, removing the projection lowers SF1 by 14.8 points under maximum-likelihood training and by 1.8 under SemInfo.
The soft-torus penalty reduces the drop only to 8.3 points under maximum-likelihood training and leaves it at 1.8 under SemInfo, suggesting that a regularizer pulling the embeddings toward the torus cannot replace the projection that keeps them exactly on it.
Under Xavier initialization, the sphere and no projection fall short of the full Hol-PCFG by 1.2 and 5.8 points under maximum-likelihood training and by 2.0 and 1.7 under SemInfo, respectively.
The full Hol-PCFG (torus initialization and projection) thus achieves the highest SF1 among the compared configurations under both objectives.%
\vspace{-0.5em}
\paragraph{Ablation Gaps under the Two Objectives.}
For the scoring-function and torus ablations, the gap to the full model is markedly smaller under SemInfo than under maximum-likelihood training,\footnote{The exception is the sphere projection, whose gap stays within 2.0 points under both objectives.} whereas the scaling-parameter ablation remains severe under both.
This is consistent with the properties of SemInfo (\S\ref{ssec:bg-seminfo}).
Maximum-likelihood training needs a refined structure of rule probabilities to maximize the marginal likelihood of a sentence, whereas SemInfo supplies a constituent-level semantic signal via REINFORCE, so a partially compromised rule-scoring operation affects SF1 less.
Even so, under \mbox{SemInfo}, HolE outperforms all the alternative scoring functions, and the benefits of torus embeddings and learnable scaling parameters remain clear.
These design choices are thus not artifacts of a single objective but structurally improve the optimization behavior of Hol-PCFG.

\begin{figure}[t]
\centering
\includegraphics[width=\linewidth]{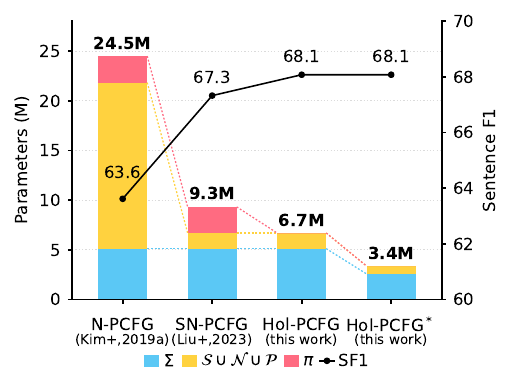}
\vspace{-3em}
\caption{Component-wise parameter counts and PTB SF1 for N-PCFG-family models in the SemInfo setting ($\Sigma$: vocabulary embeddings; $\mathcal{S}\cup \mathcal{N}\cup \mathcal{P}$: symbol-side parameters; $\pi$: production-rule scoring parameters).}
\label{fig:params_vs_f1}
\end{figure}

\subsection{Parameter Efficiency}
\label{ssec:disc-params}
As shown in Figure~\ref{fig:params_vs_f1},
Hol-PCFG outperforms N-PCFG and SN-PCFG with far fewer parameters.
In particular, production-rule scoring parameters are reduced by 99.94\% relative to SN-PCFG, and most of the parameters of Hol-PCFG are allocated to vocabulary and grammar-symbol embeddings.
This reduction arises because the asymmetric parent--child interaction that SN-PCFG learns in its MLP weights is built into circular correlation, 
so only one $d$-dimensional relation vector needs to be learned for each of the left-child, right-child, and lexical-emission roles (\S\ref{ssec:bg-hole}).
When the SN-PCFG rule scorer is shrunk to a size comparable to the Hol-PCFG scorer, the SF1 gap between SN-PCFG and Hol-PCFG becomes even larger than with the full scorer (Appendix~\ref{app:reduced-sn}).

Furthermore, the DFT of a real-valued vector satisfies Hermitian symmetry ($\hat{x}_{d-k} = \overline{\hat{x}_k}$; \S\ref{ssec:method-torus}).
The torus representation can therefore store only the phase spectrum, roughly halving the number of checkpoint parameters (``Hol-PCFG$^*$'' in Figure~\ref{fig:params_vs_f1}).
The inverse DFT expands them back into real-valued space in loading a model.%
\footnote{This halving is currently limited to checkpoint storage. Keeping rule scoring and parsing entirely in the frequency domain could also reduce memory use in training and inference.}

This parameter efficiency does not, however, allow dropping the conditional independence of the two children (\S\ref{ssec:method-grammar}).
The obstacle is not the number of scorer parameters but the size of the rule table in the inside algorithm (as discussed for N-PCFGs in \S\ref{ssec:bg-npcfg}), which is $(|\mathcal{N}|+|\mathcal{P}|)/2$ times larger without this assumption.
Even in a small grammar with a tractable joint table, dropping the assumption improves SF1 far less than scaling up the grammar size under the assumption (Appendix~\ref{app:joint-scoring}).

\subsection{Parsing Japanese without Morphological Segmentation}
\label{ssec:disc-char}
\begin{table}[t]
\centering
\footnotesize
\begin{tabular}{@{}lcc@{}}
\toprule
Method & SF1 & Morpheme recall \\
\midrule
Left Branching
  & $29.4$ & $11.8$ \\
Right Branching
  & $9.8$ & $7.5$ \\
\midrule
SN-PCFG
  & $44.7_{\pm 15.8}$ & $65.6_{\pm 4.0}$ \\
{\bf Hol-PCFG}
  & $\mathbf{58.9}_{\pm 0.7}$ & $\mathbf{74.3}_{\pm 2.3}$ \\
\bottomrule
\end{tabular}
\caption{Character-level Japanese parsing on the KTB test set (SN-PCFG and Hol-PCFG over five random seeds).
All methods produce trees over characters.}
\label{tab:char-results}
\end{table}

{\abovecaptionskip=1ex
\begin{figure}[t]
  \centering
  \includegraphics[width=\linewidth]{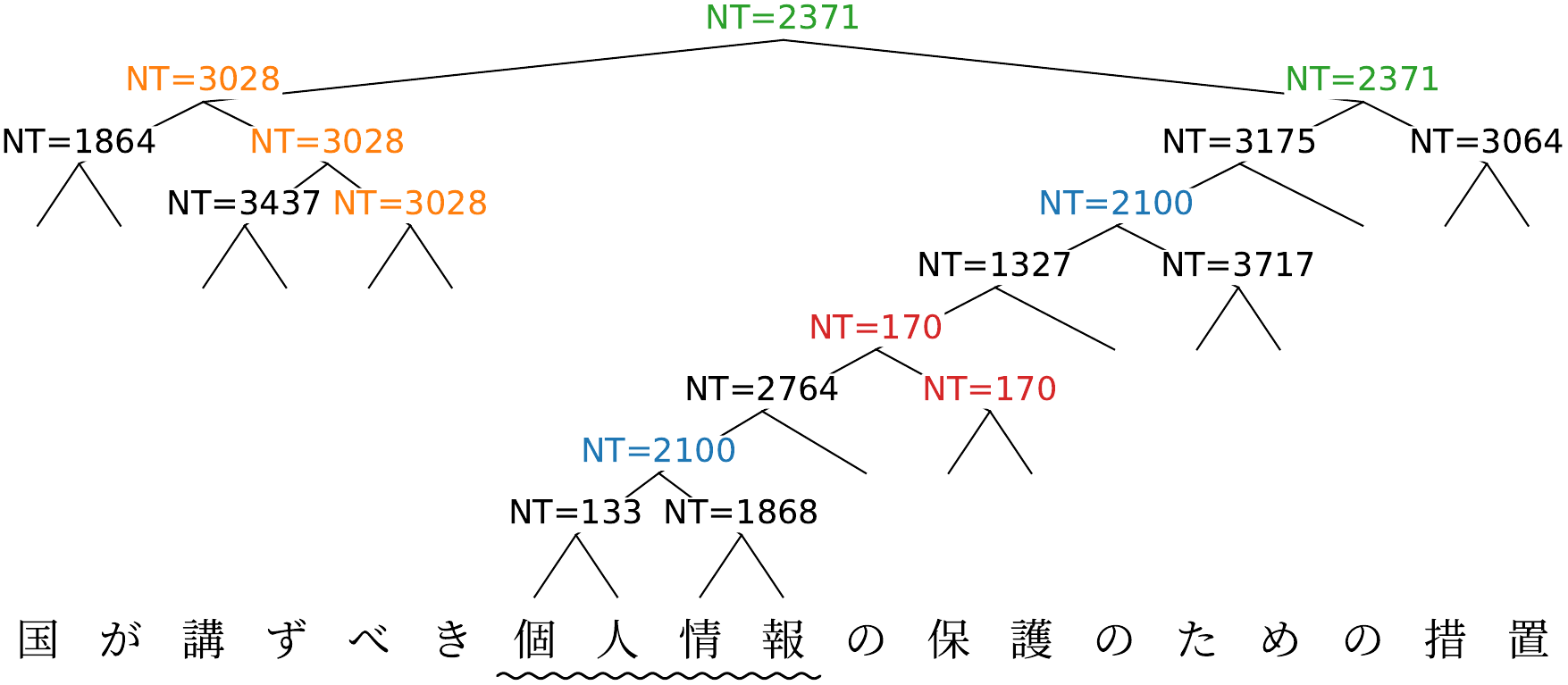}
  \caption{Character-level unsupervised constituency tree built by Hol-PCFG for a Japanese sentence meaning ``measures that the government should take to protect personal
  information.''
  See the text for the {\wave} part.
  Each color denotes one learned nonterminal.
  }
  \label{fig:char-example}
\end{figure}
}

Japanese constituency parsing is commonly performed on morpheme-segmented inputs, which fixes the granularity
of terminal symbols before inducing grammars.
To assess the generality of Hol-PCFG with respect to this granularity, we train Hol-PCFG on KTB with characters as terminal symbols and examine whether it can induce constituent structure without external segmentation.

For comparison, we train SN-PCFG on the same character sequences under the same settings as Hol-PCFG.
To evaluate constituent structures above morphemes, we keep predicted constituents that start and end on morpheme boundaries, and map them to morpheme spans to compute SF1 against the same gold morpheme-level trees as in Table~\ref{tab:results-multilingual}.
Hol-PCFG achieves an SF1 of $58.9_{\pm 0.7}$ (Table~\ref{tab:char-results}), close to the $59.5_{\pm 0.3}$ obtained by training on morphemes (Table~\ref{tab:results-multilingual}), while SN-PCFG drops from $58.1_{\pm 1.6}$ to $44.7_{\pm 15.8}$ under the same change of input.

Computed over morpheme spans, this performance does not reveal whether the model groups the characters of each morpheme into one constituent, \ie, recovers morphemes without supervision.
We therefore measure recall on morphemes as the percentage of multi-character gold morphemes in the KTB test set whose character spans are predicted constituents.
Hol-PCFG achieves a recall of $74.3_{\pm 2.3}$ (Table~\ref{tab:char-results}), 
above SN-PCFG ($65.6_{\pm 4.0}$) and far above LB (11.8) and RB (7.5).
In the qualitative example in Figure~\ref{fig:char-example}, Hol-PCFG identifies structures within the underlined compound \jpmin{個人情報} (``personal information'') by grouping the morpheme-like units \jpmin{個人} (``individual'') and \jpmin{情報} (``information'').
Overall, without a morphological analyzer, Hol-PCFG induces constituent structures 
directly from characters almost as accurately as from morpheme-segmented inputs,
and recovers morphemes more accurately than SN-PCFG.

\subsection{Parsing Non-Linguistic Data}
\label{ssec:disc-kaomoji}
\suppressfloats
\begin{table}[t]
\centering
\footnotesize
\begin{tabular}{@{}lcc@{}}
\toprule
Method & \cellbox{5em}{\centering Bracket\\[.4ex] recall}
       & \cellbox{5em}{\centering Multi-face\\[.4ex] accuracy} \\
\midrule
Left Branching
  & $39.9$ & $0.0$ \\
Right Branching
  & $26.8$ & $0.0$ \\
Linear UB
  & $79.4$ & $0.0$ \\
\midrule
SN-PCFG
  & $75.2_{\pm 20.5}$ & $56.7_{\pm 40.6}$ \\
{\bf Hol-PCFG}
  & $\mathbf{95.0}_{\pm 0.9}$ & $\mathbf{94.0}_{\pm 2.0}$ \\
\bottomrule
\end{tabular}
\caption{Results of \emph{kaomoji} parsing on the test split.
Linear UB is the upper bound for linear grammars.
SN-PCFG and Hol-PCFG use three random seeds.}
\label{tab:kaomoji-results}
\end{table}

We next examine whether Hol-PCFG generalizes to non-linguistic text-like data that are expected to carry some latent structure.
As a case study, we use \emph{kaomoji}, Japanese-style emoticons that depict faces or gestures with ASCII and Unicode characters, such as \kaomoji{(\textasciicircum\textunderscore\textasciicircum)} or \kaomoji{(T\textunderscore T)}, in which paired brackets enclose the eyes and mouth of a face.
Many \emph{kaomoji} thus exhibit an implicit compositional structure despite not being ordinary text.

We use the Emoticon/Kaomoji Dataset\footnote{\url{https://github.com/ekohrt/emoticon_kaomoji_dataset}, last accessed July 1, 2026.} of roughly 62{,}000 text emoticons and \emph{kaomoji} from online emoticon databases.
We treat each \emph{kaomoji} as a character sequence to train Hol-PCFG by maximum likelihood, and parse it with the model.

{\abovecaptionskip=1.2ex
\begin{figure}[tb]
  \centering
  \includegraphics[width=0.9\linewidth]{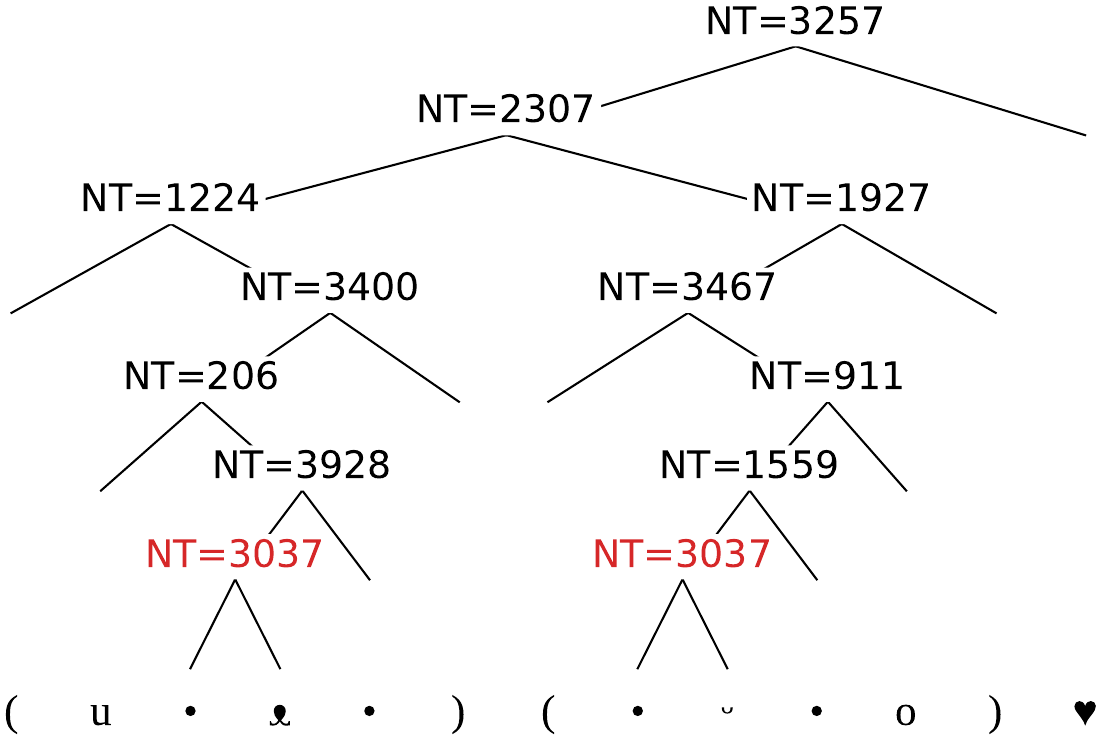}
  \caption{Hol-PCFG parse of
    ``$\vcenter{\hbox{\includegraphics[height=1em]{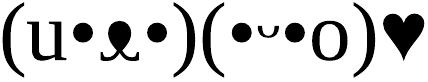}}}$'',
    a \emph{kaomoji} depicting two faces turned toward each other, followed by a heart,
    suggesting mutual affection.}
    
  \label{fig:kaomoji-example}
\end{figure}
}

Because \emph{kaomoji} have no gold trees, we measure the recovery of bracket pairs (bracket recall) and of faces as constituents (multi-face accuracy).
Bracket recall is the percentage of bracket pairs whose spans are constituents of the predicted tree, and multi-face accuracy is the percentage of \emph{kaomoji} with two or more faces (outermost bracket pairs) in which every face is a constituent.\footnote{For example, the pair in \kaomoji{\textbackslash(\textasciicircum o\textasciicircum)/} is recovered only if \kaomoji{(\textasciicircum o\textasciicircum)} is a constituent, and \kaomoji{(o\textasciicircum\textasciicircum)o(\textasciicircum\textasciicircum o)} counts only if both \kaomoji{(o\textasciicircum\textasciicircum)} and \kaomoji{(\textasciicircum\textasciicircum o)} are constituents. Pairs enclosing a whole \emph{kaomoji} are excluded from bracket recall, as they are constituents of every tree. The two metrics use subsets automatically extracted from the held-out test split of 2{,}000 \emph{kaomoji} (1{,}219 \emph{kaomoji} for bracket recall and 301 for multi-face accuracy).}

We compare Hol-PCFG both with SN-PCFG trained under the same settings and with regular and linear grammars, which are less expressive than CFGs.
A regular grammar is either right-linear (${A \to w\, B}$, ${A \to w}$) or left-linear (${A \to B\, w}$, ${A \to w}$) and thus can assign only right- or left-branching trees.
A linear grammar may mix these rule types and lies between regular grammars and CFGs in expressive power.
For linear grammars, we report an upper bound (Linear UB) obtained by selecting for each \emph{kaomoji} the tree with the most bracket pairs as constituents among the trees a linear grammar can assign.

In Table~\ref{tab:kaomoji-results}, LB and RB recover at most 39.9\% of bracket pairs, and even Linear UB only 79.4\%.
Owing to the limited expressive power of these grammars, none of them can recognize every face of a multi-face \emph{kaomoji} as a constituent.
Meanwhile, 
Hol-PCFG achieves $95.0_{\pm 0.9}$ bracket recall and $94.0_{\pm 2.0}$ multi-face accuracy, exceeding these bounds and SN-PCFG ($75.2_{\pm 20.5}$ and $56.7_{\pm 40.6}$), which exceeds Linear UB only in multi-face accuracy.
Hol-PCFG thus recovers the bracket structure of \emph{kaomoji} more accurately than SN-PCFG, a structure beyond regular and linear grammars.

In Figure~\ref{fig:kaomoji-example}, Hol-PCFG produces a plausible parse that captures salient facial structure.
This parse groups each parenthesized face into its own subtree and attaches the trailing symbol separately, consistent with a natural visual segmentation.
These results suggest that Hol-PCFG can also be used to explore latent structure in non-linguistic text-like sequences.

\subsection{Embedding Geometry and Hierarchical Organization of Induced Nonterminals}
\label{ssec:disc-hierarchy}
\begin{figure}[t]
  \centering
  \includegraphics[width=0.9\linewidth]{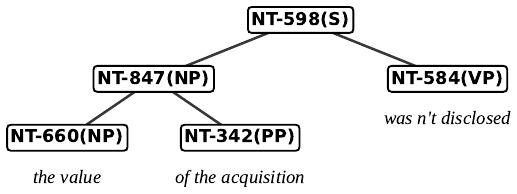}
  \includegraphics[width=\linewidth]{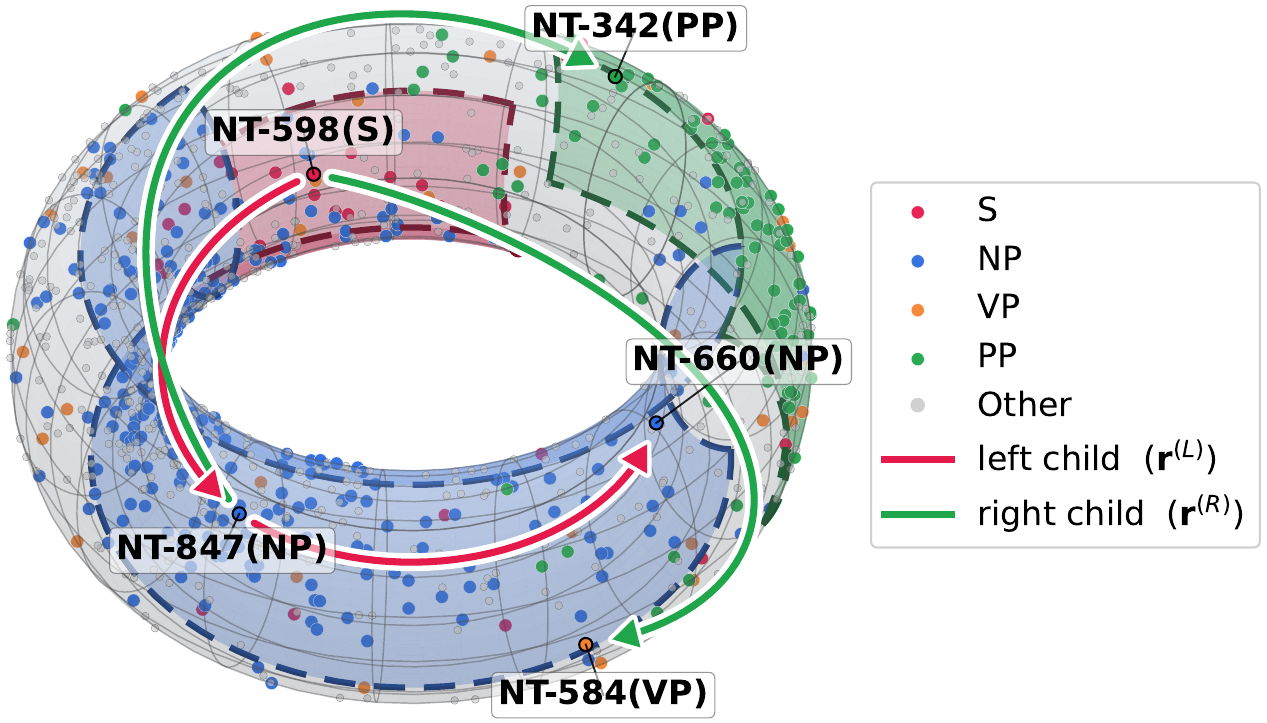}
  \caption{Visualization of the Hol-PCFG embedding space on a two-dimensional torus. The arrows indicate parent-to-child displacements in phase coordinates between the nonterminal embeddings appearing in the parse tree of the example sentence (\emph{``the value of the acquisition was n't disclosed''}).}
  \label{fig:torus-visualization}
\end{figure}
\begin{table}[t]
\centering
\scriptsize
\setlength{\tabcolsep}{4pt}
\begin{tabular}{@{}llc@{}}
\toprule
Model & Vectors compared & Agreement (\%) \\
\midrule
Hol-PCFG & Embedding $\mathbf{w}_A$          & $\mathbf{52.2}_{\pm 0.6}$ \\
SN-PCFG  & Embedding $\mathbf{w}_A$          & $31.4_{\pm 1.8}$ \\
SN-PCFG  & Parent-side vector                & $32.0_{\pm 1.6}$ \\
SN-PCFG  & Direction of parent-side vector   & $34.5_{\pm 1.5}$ \\
\bottomrule
\end{tabular}
\caption{Agreement between the ten nearest neighbors of a nonterminal in the embedding space and its ten nearest neighbors in terms of rule distributions, under maximum-likelihood training on PTB (five seeds).
For SN-PCFG, we compare the embeddings, the parent-side vectors into which its MLP transforms them before the inner product with the child-side vectors, and the directions of the parent-side vectors alone (Appendix~\ref{app:nt-hierarchy}).}
\label{tab:nn-agreement}
\end{table}
{\abovecaptionskip=3pt
\begin{figure*}[t]
  \centering
  \includegraphics[width=\textwidth]{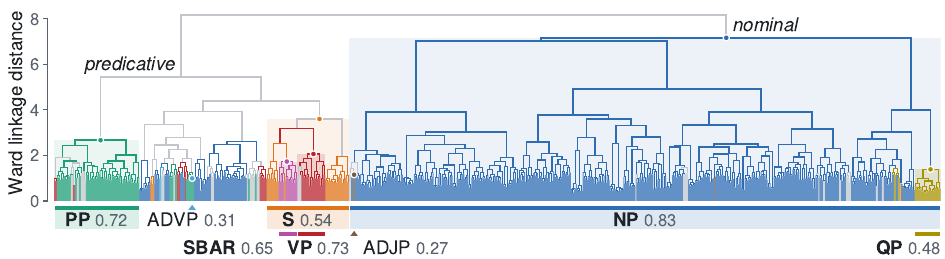}
  \caption{Dendrogram of nonterminal embeddings of Hol-PCFG trained on PTB with the
  SemInfo objective ($|\mathcal{N}| = 1024$).
  Leaves are induced nonterminals, merged bottom-up by Ward clustering with the Euclidean distance between embeddings.
  Each link is colored by the dominant gold category of the constituents
  generated by the corresponding subtree (gray if no category reaches 50\%).
  For each category, the best-matching subtree is marked by a shaded box, a
  dot on its root, and a bar (a triangle for a two-leaf subtree) labeled with
  its mass-weighted F1.}
  \label{fig:nt-hierarchy}
\end{figure*}
}

We qualitatively analyze the geometric structure that Hol-PCFG learns on the torus.
Figure~\ref{fig:torus-visualization} visualizes on a two-dimensional torus the nonterminal embeddings of Hol-PCFG trained on PTB with the SemInfo objective.
For visualization, induced spans are associated with gold constituent categories that they overlap.
We select two frequency components by Fisher discriminant ratio so that the two most frequent gold categories (NP and VP) and a third class of the remaining categories separate as clearly as possible.

Although this projection discards information, the projected embeddings tend to form clusters sharing characteristic phase patterns on the torus, 
some apparently corresponding to NP, PP, and S.

Because every rule probability in Hol-PCFG is a closed-form function of 
symbol embeddings, the learned grammar itself can be analyzed geometrically.
On the torus, the distance between the
embeddings of two symbols reflects the difference in the peak positions of
their rule distributions, not in their sharpness (\S\ref{ssec:method-torus}).
This correspondence is observed in trained models.
In Hol-PCFG under
maximum-likelihood training, on average 52.2\% of the ten nearest neighbors
of a nonterminal in the embedding space are also among its ten
nearest neighbors in terms of rule distributions.
In SN-PCFG, whose scorer leaves the vector norms unconstrained, this
agreement becomes 31.4--34.5\% (Table~\ref{tab:nn-agreement}; see Appendix~\ref{app:nt-hierarchy} for the definition).
Therefore, we build a hierarchy over the induced symbols by Ward
agglomerative clustering of their embeddings.
To interpret the hierarchy, we soft-assign each gold constituent of the
dev split to nonterminals by the model posterior,
linking every subtree to the treebank categories it
tends to realize (Appendix~\ref{app:nt-hierarchy}).

Figure~\ref{fig:nt-hierarchy} shows the dendrogram.
The induced
symbols form linguistically coherent clusters.
The first split separates a larger nominal region from a predicative
region of the verbal (VP), clausal (S, SBAR), and prepositional (PP)
symbols, yet this region contains a substantial subset of nominal
symbols.
In the predicative region, the clausal categories lie in the verbal
cluster.
S merges with VP rather than forming its own subtree, and SBAR
is nested in the VP-dominated branch, echoing the view of the clause as an
extended projection of the verb.
In the nominal region, numeric quantity phrases (QP) form a distinct
sub-cluster.
Without supervision, the grammar thus induces a coarse phrasal ontology close to that of the treebank.

\section{Conclusion}
\label{sec:conclusion}
In this paper, 
we proposed Holographic Neural PCFG (Hol-PCFG) that scores PCFG production rules algebraically as HolE-based relation scoring among grammar-symbol embeddings to capture the directed, role-dependent parent--child relations with parameter-efficient bilinear operations on a high-dimensional torus.
Hol-PCFG achieved state-of-the-art performance in six languages with far fewer parameters and higher training stability.
Further analyses showed that Hol-PCFG generalizes beyond word- and morpheme-level inputs, parsing directly from characters at nearly morpheme-level performance and inducing plausible structures 
also in non-linguistic text-like data.

These results have limitations.
From a modeling perspective, Hol-PCFG inherits from SN-PCFG the conditional independence of the two children of a binary rule given the parent, so sibling correlations are not captured.
Dropping this assumption in a small grammar yielded a far smaller performance gain than keeping it and scaling up the grammar size (\S\ref{ssec:disc-params}), but modeling the dependence between the two children in large grammars remains a challenge.
From an empirical perspective, our character-level results are so far confined to Japanese, and whether these findings generalize to languages with other morphological typologies and writing systems remains open.

For future work, adding sentence-level latent variables, as in C-PCFG and SC-PCFG, could capture sentence-specific features better and further improve performance.
Implementation-level optimizations are also important, such as improving memory efficiency and speed by 
parsing entirely in the frequency domain.

\section*{Acknowledgements}
This work was supported by JST ACT-X, Japan, Grant Number JPMJAX24CS.

\bibliography{references}

@inproceedings{chen2025improving,
  title     = {Improving Unsupervised Constituency Parsing via Maximizing Semantic Information},
  author    = {Junjie Chen and Xiangheng He and Yusuke Miyao and Danushka Bollegala},
  booktitle = {The Thirteenth International Conference on Learning Representations},
  year      = {2025},
  url       = {https://openreview.net/forum?id=qyU5s4fzLg}
}

@inproceedings{liu-etal-2023-simple,
  title     = {Simple Hardware-Efficient {PCFG}s with Independent Left and Right Productions},
  author    = {Liu, Wei  and
               Yang, Songlin  and
               Kim, Yoon  and
               Tu, Kewei},
  editor    = {Bouamor, Houda  and
               Pino, Juan  and
               Bali, Kalika},
  booktitle = {Findings of the Association for Computational Linguistics: EMNLP 2023},
  month     = dec,
  year      = {2023},
  address   = {Singapore},
  publisher = {Association for Computational Linguistics},
  url       = {https://aclanthology.org/2023.findings-emnlp.113/},
  doi       = {10.18653/v1/2023.findings-emnlp.113},
  pages     = {1662--1669}
}

@inproceedings{kim-etal-2019-compound,
  title     = {Compound Probabilistic Context-Free Grammars for Grammar Induction},
  author    = {Kim, Yoon  and
               Dyer, Chris  and
               Rush, Alexander},
  editor    = {Korhonen, Anna  and
               Traum, David  and
               M{\`a}rquez, Llu{\'i}s},
  booktitle = {Proceedings of the 57th Annual Meeting of the Association for Computational Linguistics},
  month     = jul,
  year      = {2019},
  address   = {Florence, Italy},
  publisher = {Association for Computational Linguistics},
  url       = {https://aclanthology.org/P19-1228/},
  doi       = {10.18653/v1/P19-1228},
  pages     = {2369--2385}
}

@article{bannard-etal-2009-modeling,
  author   = {Colin Bannard  and Elena Lieven  and Michael Tomasello },
  title    = {Modeling children's early grammatical knowledge},
  journal  = {Proceedings of the National Academy of Sciences},
  volume   = {106},
  number   = {41},
  pages    = {17284-17289},
  year     = {2009},
  doi      = {10.1073/pnas.0905638106},
  url      = {https://www.pnas.org/doi/abs/10.1073/pnas.0905638106},
  eprint   = {https://www.pnas.org/doi/pdf/10.1073/pnas.0905638106}
}

@inproceedings{jin-etal-2021-character-based,
  title     = {Character-based {PCFG} Induction for Modeling the Syntactic Acquisition of Morphologically Rich Languages},
  author    = {Jin, Lifeng  and
               Oh, Byung-Doh  and
               Schuler, William},
  editor    = {Moens, Marie-Francine  and
               Huang, Xuanjing  and
               Specia, Lucia  and
               Yih, Scott Wen-tau},
  booktitle = {Findings of the Association for Computational Linguistics: EMNLP 2021},
  month     = nov,
  year      = {2021},
  address   = {Punta Cana, Dominican Republic},
  publisher = {Association for Computational Linguistics},
  url       = {https://aclanthology.org/2021.findings-emnlp.371/},
  doi       = {10.18653/v1/2021.findings-emnlp.371},
  pages     = {4367--4378}
}

@article{jin-etal-2021-depth,
  title     = {Depth-Bounded Statistical {PCFG} Induction as a Model of Human Grammar Acquisition},
  author    = {Jin, Lifeng  and
               Schwartz, Lane  and
               Doshi-Velez, Finale  and
               Miller, Timothy  and
               Schuler, William},
  journal   = {Computational Linguistics},
  volume    = {47},
  number    = {1},
  month     = mar,
  year      = {2021},
  address   = {Cambridge, MA},
  publisher = {MIT Press},
  url       = {https://aclanthology.org/2021.cl-1.7/},
  doi       = {10.1162/coli_a_00399},
  pages     = {181--216}
}

@inproceedings{nickel2016holographic,
  title     = {Holographic Embeddings of Knowledge Graphs},
  author    = {Nickel, Maximilian and Rosasco, Lorenzo and Poggio, Tomaso},
  booktitle = {Proceedings of the AAAI Conference on Artificial Intelligence},
  volume    = {30},
  pages     = {1955--1961},
  year      = {2016},
  doi       = {10.1609/aaai.v30i1.10314},
  url       = {https://ojs.aaai.org/index.php/AAAI/article/view/10314}
}

@article{marcus-etal-1993-building,
  title   = {Building a Large Annotated Corpus of {E}nglish: The {P}enn {T}reebank},
  author  = {Marcus, Mitchell P.  and  Santorini, Beatrice  and  Marcinkiewicz, Mary Ann},
  journal = {Computational Linguistics},
  volume  = {19},
  number  = {2},
  pages   = {313--330},
  year    = {1993},
  url     = {https://aclanthology.org/J93-2004/}
}

@article{xue-etal-2005-penn,
  title   = {The {P}enn {C}hinese {T}ree{B}ank: Phrase Structure Annotation of a Large Corpus},
  author  = {Xue, Naiwen  and  Xia, Fei  and  Chiou, Fu-Dong  and  Palmer, Martha},
  journal = {Natural Language Engineering},
  volume  = {11},
  number  = {2},
  pages   = {207--238},
  year    = {2005},
  doi     = {10.1017/S135132490400364X}
}

@inproceedings{seddah-etal-2013-overview,
  title     = {Overview of the {SPMRL} 2013 Shared Task: A Cross-Framework Evaluation of Parsing Morphologically Rich Languages},
  author    = {Seddah, Djam{\'e}  and
               Tsarfaty, Reut  and
               K{\"u}bler, Sandra  and
               Candito, Marie  and
               Choi, Jinho D.  and
               Farkas, Rich{\'a}rd  and
               Foster, Jennifer  and
               Goenaga, Iakes  and
               Gojenola Galletebeitia, Koldo  and
               Goldberg, Yoav  and
               Green, Spence  and
               Habash, Nizar  and
               Kuhlmann, Marco  and
               Maier, Wolfgang  and
               Nivre, Joakim  and
               Przepi{\'o}rkowski, Adam  and
               Roth, Ryan  and
               Seeker, Wolfgang  and
               Versley, Yannick  and
               Vincze, Veronika  and
               Woli{\'n}ski, Marcin  and
               Wr{\'o}blewska, Alina  and
               Villemonte de la Clergerie, Eric},
  editor    = {Goldberg, Yoav  and
               Marton, Yuval  and
               Rehbein, Ines  and
               Versley, Yannick},
  booktitle = {Proceedings of the Fourth Workshop on Statistical Parsing of Morphologically-Rich Languages},
  month     = oct,
  year      = {2013},
  address   = {Seattle, Washington, USA},
  publisher = {Association for Computational Linguistics},
  url       = {https://aclanthology.org/W13-4917/},
  pages     = {146--182}
}

@article{baker-1979-trainable,
  title     = {Trainable Grammars for Speech Recognition},
  author    = {Baker, James K.},
  journal   = {The Journal of the Acoustical Society of America},
  volume    = {65},
  number    = {S1},
  pages     = {S132--S132},
  year      = {1979},
  publisher = {Acoustical Society of America},
  doi       = {10.1121/1.2017061}
}

@inproceedings{drozdov-etal-2019-unsupervised,
  title     = {Unsupervised Latent Tree Induction with Deep Inside-Outside Recursive Auto-Encoders},
  author    = {Drozdov, Andrew  and  Verga, Patrick  and  Yadav, Mohit  and  Iyyer, Mohit  and  McCallum, Andrew},
  booktitle = {Proceedings of the 2019 Conference of the North American Chapter of the Association for Computational Linguistics: Human Language Technologies, Volume 1 (Long and Short Papers)},
  month     = jun,
  year      = {2019},
  address   = {Minneapolis, Minnesota},
  publisher = {Association for Computational Linguistics},
  url       = {https://aclanthology.org/N19-1116/},
  doi       = {10.18653/v1/N19-1116},
  pages     = {1129--1141}
}

@inproceedings{drozdov-etal-2020-unsupervised,
  title     = {Unsupervised Parsing with {S}-{DIORA}: Single Tree Encoding for Deep Inside-Outside Recursive Autoencoders},
  author    = {Drozdov, Andrew  and  Rongali, Subendhu  and  Chen, Yi-Pei  and  O'Gorman, Tim  and  Iyyer, Mohit  and  McCallum, Andrew},
  booktitle = {Proceedings of the 2020 Conference on Empirical Methods in Natural Language Processing (EMNLP)},
  month     = nov,
  year      = {2020},
  address   = {Online},
  publisher = {Association for Computational Linguistics},
  url       = {https://aclanthology.org/2020.emnlp-main.392/},
  doi       = {10.18653/v1/2020.emnlp-main.392},
  pages     = {4832--4845}
}

@inproceedings{klein-manning-2002-generative,
  title     = {A Generative Constituent-Context Model for Improved Grammar Induction},
  author    = {Klein, Dan  and  Manning, Christopher D.},
  booktitle = {Proceedings of the 40th Annual Meeting of the Association for Computational Linguistics},
  month     = jul,
  year      = {2002},
  address   = {Philadelphia, Pennsylvania, USA},
  publisher = {Association for Computational Linguistics},
  url       = {https://aclanthology.org/P02-1017/},
  doi       = {10.3115/1073083.1073106},
  pages     = {128--135}
}

@inproceedings{seginer-2007-fast,
  title     = {Fast Unsupervised Incremental Parsing},
  author    = {Seginer, Yoav},
  booktitle = {Proceedings of the 45th Annual Meeting of the Association of Computational Linguistics},
  month     = jun,
  year      = {2007},
  address   = {Prague, Czech Republic},
  publisher = {Association for Computational Linguistics},
  url       = {https://aclanthology.org/P07-1049/},
  pages     = {384--391}
}

@inproceedings{yang-etal-2021-pcfgs,
  title     = {{PCFG}s Can Do Better: Inducing Probabilistic Context-Free Grammars with Many Symbols},
  author    = {Yang, Songlin  and  Zhao, Yanpeng  and  Tu, Kewei},
  booktitle = {Proceedings of the 2021 Conference of the North American Chapter of the Association for Computational Linguistics: Human Language Technologies},
  month     = jun,
  year      = {2021},
  address   = {Online},
  publisher = {Association for Computational Linguistics},
  url       = {https://aclanthology.org/2021.naacl-main.117/},
  doi       = {10.18653/v1/2021.naacl-main.117},
  pages     = {1487--1498}
}

@article{plate1995holographic,
  title   = {Holographic Reduced Representations},
  author  = {Plate, Tony A.},
  journal = {IEEE Transactions on Neural Networks},
  volume  = {6},
  number  = {3},
  pages   = {623--641},
  year    = {1995},
  doi     = {10.1109/72.377968}
}

@inproceedings{yamaki-etal-2023-holographic,
  title     = {Holographic {CCG} Parsing},
  author    = {Yamaki, Ryosuke  and  Taniguchi, Tadahiro  and  Mochihashi, Daichi},
  booktitle = {Proceedings of the 61st Annual Meeting of the Association for Computational Linguistics (Volume 1: Long Papers)},
  month     = jul,
  year      = {2023},
  address   = {Toronto, Canada},
  publisher = {Association for Computational Linguistics},
  url       = {https://aclanthology.org/2023.acl-long.15/},
  doi       = {10.18653/v1/2023.acl-long.15},
  pages     = {262--276}
}

@article{aizawa2003information,
  title   = {An Information-theoretic Perspective of {tf}-{idf} Measures},
  author  = {Aizawa, Akiko},
  journal = {Information Processing \& Management},
  volume  = {39},
  number  = {1},
  pages   = {45--65},
  year    = {2003},
  doi     = {10.1016/S0306-4573(02)00021-3}
}

@inproceedings{kim-etal-2019-unsupervised,
  title     = {Unsupervised Recurrent Neural Network Grammars},
  author    = {Kim, Yoon  and  Rush, Alexander  and  Yu, Lei  and  Kuncoro, Adhiguna  and  Dyer, Chris  and  Melis, G{\'a}bor},
  booktitle = {Proceedings of the 2019 Conference of the North American Chapter of the Association for Computational Linguistics: Human Language Technologies, Volume 1 (Long and Short Papers)},
  month     = jun,
  year      = {2019},
  address   = {Minneapolis, Minnesota},
  publisher = {Association for Computational Linguistics},
  url       = {https://aclanthology.org/N19-1114/},
  doi       = {10.18653/v1/N19-1114},
  pages     = {1105--1117}
}

@inproceedings{stern-etal-2017-minimal,
  title     = {A Minimal Span-Based Neural Constituency Parser},
  author    = {Stern, Mitchell  and  Andreas, Jacob  and  Klein, Dan},
  booktitle = {Proceedings of the 55th Annual Meeting of the Association for Computational Linguistics (Volume 1: Long Papers)},
  month     = jul,
  year      = {2017},
  address   = {Vancouver, Canada},
  publisher = {Association for Computational Linguistics},
  url       = {https://aclanthology.org/P17-1076/},
  doi       = {10.18653/v1/P17-1076},
  pages     = {818--827}
}

@inproceedings{zhao-titov-2020-visually,
  title     = {Visually Grounded Compound {PCFG}s},
  author    = {Zhao, Yanpeng  and  Titov, Ivan},
  booktitle = {Proceedings of the 2020 Conference on Empirical Methods in Natural Language Processing (EMNLP)},
  month     = nov,
  year      = {2020},
  address   = {Online},
  publisher = {Association for Computational Linguistics},
  url       = {https://aclanthology.org/2020.emnlp-main.354/},
  doi       = {10.18653/v1/2020.emnlp-main.354},
  pages     = {4369--4379}
}

@article{zhu-etal-2020-return,
  title   = {The Return of Lexical Dependencies: Neural Lexicalized {PCFG}s},
  author  = {Zhu, Hao  and  Bisk, Yonatan  and  Neubig, Graham},
  journal = {Transactions of the Association for Computational Linguistics},
  volume  = {8},
  pages   = {647--661},
  year    = {2020},
  url     = {https://aclanthology.org/2020.tacl-1.42/},
  doi     = {10.1162/tacl_a_00337}
}

@inproceedings{yang-etal-2021-neural,
  title     = {Neural Bi-Lexicalized {PCFG} Induction},
  author    = {Yang, Songlin  and  Zhao, Yanpeng  and  Tu, Kewei},
  booktitle = {Proceedings of the 59th Annual Meeting of the Association for Computational Linguistics and the 11th International Joint Conference on Natural Language Processing (Volume 1: Long Papers)},
  month     = aug,
  year      = {2021},
  address   = {Online},
  publisher = {Association for Computational Linguistics},
  url       = {https://aclanthology.org/2021.acl-long.209/},
  doi       = {10.18653/v1/2021.acl-long.209},
  pages     = {2688--2699}
}

@inproceedings{yang-etal-2023-unsupervised,
  title     = {Unsupervised Discontinuous Constituency Parsing with Mildly Context-Sensitive Grammars},
  author    = {Yang, Songlin  and  Levy, Roger  and  Kim, Yoon},
  booktitle = {Proceedings of the 61st Annual Meeting of the Association for Computational Linguistics (Volume 1: Long Papers)},
  month     = jul,
  year      = {2023},
  address   = {Toronto, Canada},
  publisher = {Association for Computational Linguistics},
  url       = {https://aclanthology.org/2023.acl-long.316/},
  doi       = {10.18653/v1/2023.acl-long.316},
  pages     = {5747--5766}
}

@inproceedings{cao-etal-2020-unsupervised,
  title     = {Unsupervised Parsing via Constituency Tests},
  author    = {Cao, Steven  and  Kitaev, Nikita  and  Klein, Dan},
  editor    = {Webber, Bonnie  and  Cohn, Trevor  and  He, Yulan  and  Liu, Yang},
  booktitle = {Proceedings of the 2020 Conference on Empirical Methods in Natural Language Processing (EMNLP)},
  month     = nov,
  year      = {2020},
  address   = {Online},
  publisher = {Association for Computational Linguistics},
  url       = {https://aclanthology.org/2020.emnlp-main.389/},
  doi       = {10.18653/v1/2020.emnlp-main.389},
  pages     = {4798--4808}
}

@inproceedings{chen-etal-2024-unsupervised,
  title     = {Unsupervised Parsing by Searching for Frequent Word Sequences among Sentences with Equivalent Predicate-Argument Structures},
  author    = {Chen, Junjie  and  He, Xiangheng  and  Bollegala, Danushka  and  Miyao, Yusuke},
  editor    = {Ku, Lun-Wei  and  Martins, Andre  and  Srikumar, Vivek},
  booktitle = {Findings of the Association for Computational Linguistics: ACL 2024},
  month     = aug,
  year      = {2024},
  address   = {Bangkok, Thailand},
  publisher = {Association for Computational Linguistics},
  url       = {https://aclanthology.org/2024.findings-acl.225},
  doi       = {10.18653/v1/2024.findings-acl.225},
  pages     = {3760--3772}
}

@inproceedings{akiba2019optuna,
  author    = {Akiba, Takuya and Sano, Shotaro and Yanase, Toshihiko and Ohta, Takeru and Koyama, Masanori},
  title     = {Optuna: A Next-generation Hyperparameter Optimization Framework},
  year      = {2019},
  isbn      = {9781450362016},
  publisher = {Association for Computing Machinery},
  address   = {New York, NY, USA},
  url       = {https://doi.org/10.1145/3292500.3330701},
  doi       = {10.1145/3292500.3330701},
  booktitle = {Proceedings of the 25th ACM SIGKDD International Conference on Knowledge Discovery \& Data Mining},
  pages     = {2623--2631},
  numpages  = {9},
  location  = {Anchorage, AK, USA},
  series    = {KDD '19}
}

@inproceedings{shen-etal-2018-neural,
  title     = {Neural Language Modeling by Jointly Learning Syntax and Lexicon},
  author    = {Shen, Yikang  and  Lin, Zhouhan  and  Huang, Chin-Wei  and  Courville, Aaron},
  booktitle = {International Conference on Learning Representations},
  year      = {2018},
  url       = {https://openreview.net/forum?id=rkgOLb-0W}
}

@inproceedings{shen-etal-2019-ordered,
  title     = {Ordered Neurons: Integrating Tree Structures into Recurrent Neural Networks},
  author    = {Shen, Yikang  and  Tan, Shawn  and  Sordoni, Alessandro  and  Courville, Aaron},
  booktitle = {International Conference on Learning Representations},
  year      = {2019},
  url       = {https://openreview.net/forum?id=B1l6qiR5F7}
}

@inproceedings{li-lu-2023-contextual,
  title     = {Contextual Distortion Reveals Constituency: Masked Language Models are Implicit Parsers},
  author    = {Li, Jiaxi  and  Lu, Wei},
  booktitle = {Proceedings of the 61st Annual Meeting of the Association for Computational Linguistics (Volume 1: Long Papers)},
  month     = jul,
  year      = {2023},
  address   = {Toronto, Canada},
  publisher = {Association for Computational Linguistics},
  url       = {https://aclanthology.org/2023.acl-long.285/},
  doi       = {10.18653/v1/2023.acl-long.285},
  pages     = {5208--5222}
}

@inproceedings{yang2022dynamic,
    title = "Dynamic Programming in Rank Space: Scaling Structured Inference with Low-Rank {HMM}s and {PCFG}s",
    author = "Yang, Songlin  and
      Liu, Wei  and
      Tu, Kewei",
    editor = "Carpuat, Marine  and
      de Marneffe, Marie-Catherine  and
      Meza Ruiz, Ivan Vladimir",
    booktitle = "Proceedings of the 2022 Conference of the North American Chapter of the Association for Computational Linguistics: Human Language Technologies",
    month = jul,
    year = "2022",
    address = "Seattle, United States",
    publisher = "Association for Computational Linguistics",
    url = "https://aclanthology.org/2022.naacl-main.353/",
    doi = "10.18653/v1/2022.naacl-main.353",
    pages = "4797--4809"
}

@inproceedings{shen-etal-2021-structformer,
  title     = {{S}truct{F}ormer: Joint Unsupervised Induction of Dependency and Constituency Structure from Masked Language Modeling},
  author    = {Shen, Yikang  and
               Tay, Yi  and
               Zheng, Che  and
               Bahri, Dara  and
               Metzler, Donald  and
               Courville, Aaron},
  editor    = {Zong, Chengqing  and
               Xia, Fei  and
               Li, Wenjie  and
               Navigli, Roberto},
  booktitle = {Proceedings of the 59th Annual Meeting of the Association for Computational Linguistics and the 11th International Joint Conference on Natural Language Processing (Volume 1: Long Papers)},
  month     = aug,
  year      = {2021},
  address   = {Online},
  publisher = {Association for Computational Linguistics},
  url       = {https://aclanthology.org/2021.acl-long.559/},
  doi       = {10.18653/v1/2021.acl-long.559},
  pages     = {7196--7209}
}

@inproceedings{hu-etal-2022-fast,
  title     = {Fast-{R}2{D}2: A Pretrained Recursive Neural Network based on Pruned {CKY} for Grammar Induction and Text Representation},
  author    = {Hu, Xiang  and
               Mi, Haitao  and
               Li, Liang  and
               de Melo, Gerard},
  editor    = {Goldberg, Yoav  and
               Kozareva, Zornitsa  and
               Zhang, Yue},
  booktitle = {Proceedings of the 2022 Conference on Empirical Methods in Natural Language Processing},
  month     = dec,
  year      = {2022},
  address   = {Abu Dhabi, United Arab Emirates},
  publisher = {Association for Computational Linguistics},
  url       = {https://aclanthology.org/2022.emnlp-main.181/},
  doi       = {10.18653/v1/2022.emnlp-main.181},
  pages     = {2809--2821}
}

@inproceedings{butler2012keyaki,
  title     = {{Keyaki Treebank}: Phrase structure with functional information for {Japanese}},
  author    = {Butler, Alastair and Hotta, Tomoko and Otomo, Ruriko and Yoshimoto, Kei and Zhou, Zhen and Zhu, Hong},
  booktitle = {Proceedings of Text Annotation Workshop},
  year      = {2012}
}

@inproceedings{park-kim-2025-probability,
  title     = {Probability Distribution Collapse: A Critical Bottleneck to Compact Unsupervised Neural Grammar Induction},
  author    = {Park, Jinwook  and
               Kim, Kangil},
  editor    = {Christodoulopoulos, Christos  and
               Chakraborty, Tanmoy  and
               Rose, Carolyn  and
               Peng, Violet},
  booktitle = {Proceedings of the 2025 Conference on Empirical Methods in Natural Language Processing},
  month     = nov,
  year      = {2025},
  address   = {Suzhou, China},
  publisher = {Association for Computational Linguistics},
  url       = {https://aclanthology.org/2025.emnlp-main.1694/},
  doi       = {10.18653/v1/2025.emnlp-main.1694},
  pages     = {33392--33403},
  isbn      = {979-8-89176-332-6}
}

@inproceedings{glorot2010understanding,
  title     = {Understanding the Difficulty of Training Deep Feedforward Neural Networks},
  author    = {Glorot, Xavier and Bengio, Yoshua},
  booktitle = {Proceedings of the Thirteenth International Conference on Artificial Intelligence and Statistics},
  series    = {Proceedings of Machine Learning Research},
  volume    = {9},
  pages     = {249--256},
  year      = {2010},
  url       = {https://proceedings.mlr.press/v9/glorot10a.html}
}

@inproceedings{zanzotto2012distributed,
  title     = {Distributed Tree Kernels},
  author    = {Zanzotto, Fabio Massimo and Dell'Arciprete, Lorenzo},
  booktitle = {Proceedings of the 29th International Conference on Machine Learning},
  pages     = {115--122},
  year      = {2012},
  address   = {Edinburgh, Scotland, UK},
  publisher = {Omnipress},
  isbn      = {9781450312851}
}

@inproceedings{nickel2011three,
  title     = {A Three-Way Model for Collective Learning on Multi-Relational Data},
  author    = {Nickel, Maximilian and Tresp, Volker and Kriegel, Hans-Peter},
  booktitle = {Proceedings of the 28th International Conference on Machine Learning},
  pages     = {809--816},
  year      = {2011}
}

@inproceedings{bordes2013translating,
  title     = {Translating Embeddings for Modeling Multi-relational Data},
  author    = {Bordes, Antoine and Usunier, Nicolas and Garc{\'i}a-Dur{\'a}n, Alberto and Weston, Jason and Yakhnenko, Oksana},
  booktitle = {Advances in Neural Information Processing Systems},
  volume    = {26},
  pages     = {2787--2795},
  year      = {2013},
  url       = {https://proceedings.neurips.cc/paper_files/paper/2013/hash/1cecc7a77928ca8133fa24680a88d2f9-Abstract.html}
}

@inproceedings{yang2015embedding,
  title     = {Embedding Entities and Relations for Learning and Inference in Knowledge Bases},
  author    = {Yang, Bishan and Yih, Wen-tau and He, Xiaodong and Gao, Jianfeng and Deng, Li},
  booktitle = {Proceedings of the 3rd International Conference on Learning Representations},
  year      = {2015},
  url       = {https://arxiv.org/abs/1412.6575}
}

@inproceedings{trouillon2016complex,
  title     = {Complex Embeddings for Simple Link Prediction},
  author    = {Trouillon, Th{\'e}o and Welbl, Johannes and Riedel, Sebastian and Gaussier, {\'E}ric and Bouchard, Guillaume},
  booktitle = {Proceedings of the 33rd International Conference on Machine Learning},
  series    = {Proceedings of Machine Learning Research},
  volume    = {48},
  pages     = {2071--2080},
  year      = {2016},
  url       = {https://proceedings.mlr.press/v48/trouillon16.html}
}

@inproceedings{hayashi-shimbo-2017-equivalence,
  title     = {On the Equivalence of Holographic and Complex Embeddings for Link Prediction},
  author    = {Hayashi, Katsuhiko and Shimbo, Masashi},
  booktitle = {Proceedings of the 55th Annual Meeting of the Association for Computational Linguistics (Volume 2: Short Papers)},
  pages     = {554--559},
  year      = {2017},
  address   = {Vancouver, Canada},
  publisher = {Association for Computational Linguistics},
  doi       = {10.18653/v1/P17-2088},
  url       = {https://aclanthology.org/P17-2088/}
}

@inproceedings{lacroix2018canonical,
  title     = {Canonical Tensor Decomposition for Knowledge Base Completion},
  author    = {Lacroix, Timoth{\'e}e and Usunier, Nicolas and Obozinski, Guillaume},
  booktitle = {Proceedings of the 35th International Conference on Machine Learning},
  series    = {Proceedings of Machine Learning Research},
  volume    = {80},
  pages     = {2863--2872},
  year      = {2018},
  url       = {https://proceedings.mlr.press/v80/lacroix18a.html}
}

@inproceedings{balazevic2019multi,
  title     = {Multi-relational {P}oincar{\'e} Graph Embeddings},
  author    = {Bala{\v{z}}evi{\'c}, Ivana and Allen, Carl and Hospedales, Timothy},
  booktitle = {Advances in Neural Information Processing Systems},
  volume    = {32},
  year      = {2019},
  url       = {https://proceedings.neurips.cc/paper_files/paper/2019/hash/f8b932c70d0b2e6bf071729a4fa68dfc-Abstract.html}
}
\bibliographystyle{acl_natbib}

\clearpage
\appendix
\section{Derivation and Computational Cost of Efficient Scoring}
\label{app:efficient-scoring}
This appendix derives Equation~\eqref{eq:corr_conv_identity} and compares the cost of computing the binary-rule scores with and without Efficient Scoring (\S\ref{sssec:method-efficient}).

\paragraph{Derivation of Equation~\eqref{eq:corr_conv_identity}.}
By the definition of circular correlation in Equation~\eqref{eq:circular-correlation},
\begin{equation*}
  \langle \mathbf{r},\, \mathbf{a} \star \mathbf{b} \rangle
  = \sum_{k=0}^{d-1} r_k \sum_{i=0}^{d-1} a_i\, b_{(k+i) \bmod d} .
\end{equation*}
For fixed $i$, as $k$ runs over $0, \dots, d-1$, the index $j = (k+i) \bmod d$ takes each value $0, \dots, d-1$ exactly once, and $k = (j-i) \bmod d$, so summing over $j$ instead of $k$ and exchanging the sums gives
\begin{equation*}
  \langle \mathbf{r},\, \mathbf{a} \star \mathbf{b} \rangle
  = \sum_{j=0}^{d-1} b_j \sum_{i=0}^{d-1} r_{(j-i) \bmod d}\, a_i .
\end{equation*}
The definition of circular convolution in Equation~\eqref{eq:circ_conv_def} is $(\mathbf{r} * \mathbf{a})_j = \sum_{m=0}^{d-1} r_m\, a_{(j-m) \bmod d}$, and substituting $i = (j-m) \bmod d$ likewise shows that the inner sum above equals $(\mathbf{r} * \mathbf{a})_j$.
Hence ${\langle \mathbf{r}, \mathbf{a} \star \mathbf{b} \rangle} = {\sum_{j=0}^{d-1} b_j\, (\mathbf{r} * \mathbf{a})_j} = {\langle \mathbf{b}, \mathbf{r} * \mathbf{a} \rangle}$.
This derivation uses only the definitions of circular correlation and circular convolution in Equations~\eqref{eq:circular-correlation} and~\eqref{eq:circ_conv_def}, so the identity holds for arbitrary real vectors.

\paragraph{Computational Cost.}
The quantities to compute are the HolE scores inside the exponentials of Equations~\eqref{eq:left_child_prob} and~\eqref{eq:right_child_prob} (that is, the logits before multiplication by $\tau$) for all parents $A \in \mathcal{N}$ and all children $B, C \in \mathcal{N} \cup \mathcal{P}$.
For the left child, the identity in Equation~\eqref{eq:corr_conv_identity} gives two ways to write the score: ${\langle \mathbf{r}^{(L)}, \mathbf{w}_A \star \mathbf{w}_B \rangle} = {\langle \mathbf{w}_B, \mathbf{r}^{(L)} * \mathbf{w}_A \rangle}$.
The naive computation evaluates the left-hand side by computing $\mathbf{w}_A \star \mathbf{w}_B$ and taking its inner product with $\mathbf{r}^{(L)}$ for every pair of $A \in \mathcal{N}$ and $B \in \mathcal{N} \cup \mathcal{P}$.
Efficient Scoring evaluates the right-hand side by computing $\mathbf{r}^{(L)} * \mathbf{w}_A$ for every $A \in \mathcal{N}$ and then taking its inner product with $\mathbf{w}_B$.
The right-child scores are computed likewise with $\mathbf{r}^{(R)}$.

Table~\ref{tab:efficient-scoring} counts the operations and the memory (in number of elements) of the two computations, with both $\star$ and $*$ computed by FFT.
Efficient Scoring reduces the numbers of element-wise products and inverse FFTs from ${|\mathcal{N}|(|\mathcal{N}|+|\mathcal{P}|)}$ to $|\mathcal{N}|$, and the size of the intermediate tensor that must be materialized in memory from ${|\mathcal{N}|(|\mathcal{N}|+|\mathcal{P}|)\,d}$ to $|\mathcal{N}|\,d$ elements, both by a factor of ${|\mathcal{N}|+|\mathcal{P}|}$ (12{,}288 in the maximum-likelihood configuration of \S\ref{ssec:exp-setup}).

\begin{table}[t]
\centering
\scriptsize
\setlength{\tabcolsep}{1.5pt}
\begin{tabular}{@{}lccl@{}}
\toprule
 & Naive & Efficient Scoring & Ratio \\
 & $\langle \mathbf{r}^{(L)}{,}\mathbf{w}_A {\star} \mathbf{w}_B \rangle$ & $\langle \mathbf{w}_B{,}\mathbf{r}^{(L)} {*} \mathbf{w}_A \rangle$ & \\
\midrule
\multicolumn{4}{@{}l}{\textit{Number of operations}} \\
Length-$d$ FFT
  & $|\mathcal{N}|{+}|\mathcal{P}|$
  & $|\mathcal{N}|{+}1$
  & $\approx 3.0$ \\
Element-wise product
  & $|\mathcal{N}|(|\mathcal{N}|{+}|\mathcal{P}|)$
  & $|\mathcal{N}|$
  & $12{,}288$ \\
Length-$d$ inverse FFT
  & $|\mathcal{N}|(|\mathcal{N}|{+}|\mathcal{P}|)$
  & $|\mathcal{N}|$
  & $12{,}288$ \\
Length-$d$ inner product
  & $|\mathcal{N}|(|\mathcal{N}|{+}|\mathcal{P}|)$
  & $|\mathcal{N}|(|\mathcal{N}|{+}|\mathcal{P}|)$
  & $1$ \\
\midrule
\multicolumn{4}{@{}l}{\textit{Memory (number of elements)}} \\
Inputs ($\mathbf{w}_A{,}\mathbf{w}_B{,}\mathbf{r}^{(L)}$)
  & $(|\mathcal{N}|{+}|\mathcal{P}|)d{+}d$
  & $(|\mathcal{N}|{+}|\mathcal{P}|)d{+}d$
  & $1$ \\
Intermediate tensor
  & $|\mathcal{N}|(|\mathcal{N}|{+}|\mathcal{P}|)d$
  & $|\mathcal{N}|d$
  & $12{,}288$ \\
Output scores
  & $|\mathcal{N}|(|\mathcal{N}|{+}|\mathcal{P}|)$
  & $|\mathcal{N}|(|\mathcal{N}|{+}|\mathcal{P}|)$
  & $1$ \\
\bottomrule
\end{tabular}
\caption{Operation counts and memory (in elements) of the naive computation and Efficient Scoring for the left-child HolE scores of all parent--child pairs. Ratio is Naive / Efficient Scoring in the maximum-likelihood configuration ($|\mathcal{N}| = 4096$, $|\mathcal{P}| = 8192$, $d = 512$).}
\label{tab:efficient-scoring}
\end{table}

\section{Embeddings on a Torus: Background and Derivations}
\label{app:torus}
Supplementing \S\ref{ssec:method-torus}, this appendix summarizes the facts about the discrete Fourier transform (DFT), describes the structure of $\mathbb{T}^d$, and derives the properties of $\mathbb{T}^d$ as well as the separation of the peak position and the sharpness.

\paragraph{Discrete Fourier Transform.}
Equation~\eqref{eq:torus} defines $\mathbb{T}^d$ through the DFT $\hat{\mathbf{x}} = \mathcal{F}[\mathbf{x}] \in \mathbb{C}^d$ of a vector $\mathbf{x} \in \mathbb{R}^d$.
The DFT and its inverse are
\begin{align*}
  \hat{x}_k &= \sum_{n=0}^{d-1} x_n\, e^{-2\pi i k n / d}, \\
  x_n &= \frac{1}{d} \sum_{k=0}^{d-1} \hat{x}_k\, e^{2\pi i k n / d}.
\end{align*}
We call $\hat{x}_k$ the $k$-th frequency component of $\mathbf{x}$, with amplitude $|\hat{x}_k|$ and phase $\phi_k$, so that $\hat{x}_k = |\hat{x}_k|\, e^{i\phi_k}$.
The derivations in this appendix use four facts.
See \citet{hayashi-shimbo-2017-equivalence} for circular correlation and HolE in the frequency domain.
\begin{itemize}
  \item[(i)] \emph{Hermitian symmetry.} Because $\mathbf{x}$ is real, $\overline{\hat{x}_k} = \sum_n x_n\, e^{2\pi i k n / d} = \hat{x}_{d-k}$ for $k = 1, \dots, d-1$.
  \item[(ii)] \emph{Real-valued components.} The sum $\hat{x}_0 = \sum_n x_n$ of the elements of $\mathbf{x}$ and, for even $d$, their alternating sum $\hat{x}_{d/2} = \sum_n (-1)^n x_n$ are real (the DC and Nyquist components in signal processing). By fact (i), the other components form complex-conjugate pairs.
  \item[(iii)] \emph{Parseval's theorem.} $\langle \mathbf{a}, \mathbf{b} \rangle = \frac{1}{d} \sum_k \overline{\hat{a}_k}\, \hat{b}_k$ for $\mathbf{a}, \mathbf{b} \in \mathbb{R}^d$. In particular, $\lVert \mathbf{x} \rVert^2 = \frac{1}{d} \sum_k |\hat{x}_k|^2$.
  \item[(iv)] \emph{Correlation and convolution theorems.} Circular correlation (Equation~\eqref{eq:circular-correlation}) and circular convolution (Equation~\eqref{eq:circ_conv_def}) are element-wise products in the frequency domain, $\widehat{(\mathbf{a} \star \mathbf{b})}_k = \overline{\hat{a}_k}\, \hat{b}_k$ and $\widehat{(\mathbf{a} * \mathbf{b})}_k = \hat{a}_k\, \hat{b}_k$.
\end{itemize}

\paragraph{Structure of $\mathbb{T}^d$.}
Let $\mathbf{x} \in \mathbb{T}^d$.
By Equation~\eqref{eq:torus}, $\mathbf{x}$ is real and $|\hat{x}_k| = 1$ for all $k$.
By fact (ii), $\hat{x}_0$ and, when $d$ is even, $\hat{x}_{d/2}$ are real with absolute value 1, so each of them is $+1$ or $-1$.
By fact (i), the other components are determined by the phases $\phi_1, \dots, \phi_K$ of $\hat{x}_k = e^{i\phi_k}$ with $K = \lfloor (d-1)/2 \rfloor$, since $\hat{x}_{d-k} = e^{-i\phi_k}$.
Each phase is an angle (a point on a circle), and the $K$ phases can be chosen independently.
For each choice of the signs of the real-valued components, such vectors therefore form a product of $K$ circles, that is, a $K$-dimensional torus.
$\mathbb{T}^d$ is the union of these tori, one for each choice of the signs (two for odd $d$, where only $\hat{x}_0$ is real-valued, and four for even $d$, where both $\hat{x}_0$ and $\hat{x}_{d/2}$ are).
We refer to the whole set as the torus.
Substituting these components into the inverse DFT shows that, apart from the real-valued components, every vector on $\mathbb{T}^d$ is a sum of cosine waves of frequencies $k = 1, \dots, K$ with the same amplitude.
Vectors on $\mathbb{T}^d$ differ only in the phases $\phi_k$ of these waves and in the signs of the real-valued components.

\paragraph{Properties of Vectors on $\mathbb{T}^d$.}
Three properties of the vectors on $\mathbb{T}^d$ used in \S\ref{ssec:method-torus} follow from facts (i)--(iv).
First, by fact (iii), every $\mathbf{x} \in \mathbb{T}^d$ has unit norm, as $\lVert \mathbf{x} \rVert^2 = \frac{1}{d} \sum_k |\hat{x}_k|^2 = 1$.
Second, let $\mathbf{a}, \mathbf{b} \in \mathbb{T}^d$, so that $|\hat{a}_k| = |\hat{b}_k| = 1$ for all $k$, and consider their circular correlation $\mathbf{a} \star \mathbf{b}$ (Equation~\eqref{eq:circular-correlation}) and circular convolution $\mathbf{a} * \mathbf{b}$ (Equation~\eqref{eq:circ_conv_def}).
By fact (iv), their $k$-th frequency components $\overline{\hat{a}_k}\, \hat{b}_k$ and $\hat{a}_k\, \hat{b}_k$ both have amplitude $|\hat{a}_k|\, |\hat{b}_k| = 1$.
Both vectors are also real, because their elements are sums of products of elements of $\mathbf{a}$ and $\mathbf{b}$.
Hence $\mathbf{a} \star \mathbf{b},\, \mathbf{a} * \mathbf{b} \in \mathbb{T}^d$ by Equation~\eqref{eq:torus}.
That is, $\mathbb{T}^d$ is closed under $\star$ and~$*$.
In particular, the template $\mathbf{r}^{(L)} * \mathbf{w}_A$ (\S\ref{sssec:method-efficient}) lies on $\mathbb{T}^d$ because $\mathbf{r}^{(L)}$ and $\mathbf{w}_A$ do (\S\ref{sssec:method-params}), and thus has unit norm by the first property.
Third, let $\mathbf{r} \in \mathbb{T}^d$ and $\mathbf{a}, \mathbf{b} \in \mathbb{R}^d$.
Circular convolution (Equation~\eqref{eq:circ_conv_def}) is linear in its second argument, so $\mathbf{r} * \mathbf{a} - \mathbf{r} * \mathbf{b} = \mathbf{r} * (\mathbf{a} - \mathbf{b})$, whose $k$-th frequency component is $\hat{r}_k (\hat{a}_k - \hat{b}_k)$ by fact (iv).
Since $|\hat{r}_k| = 1$, fact (iii) gives
\begin{equation*}
\begin{aligned}
  &\lVert \mathbf{r} * \mathbf{a} - \mathbf{r} * \mathbf{b} \rVert^2 \\
  &\quad = \frac{1}{d} \sum_{k=0}^{d-1} |\hat{r}_k|^2\, |\hat{a}_k - \hat{b}_k|^2 \\
  &\quad = \frac{1}{d} \sum_{k=0}^{d-1} |\hat{a}_k - \hat{b}_k|^2
   = \lVert \mathbf{a} - \mathbf{b} \rVert^2 .
\end{aligned}
\end{equation*}
Hence circular convolution with a fixed vector on $\mathbb{T}^d$ preserves distances between vectors, that is, it is an isometry.
In particular, since the relation vectors $\mathbf{r}^{(L)}$ and $\mathbf{r}^{(R)}$ lie on $\mathbb{T}^d$ (\S\ref{sssec:method-params}), the distance between the embeddings $\mathbf{w}_A$ and $\mathbf{w}_{A'}$ of two symbols equals the distance between their left-child templates $\mathbf{r}^{(L)} * \mathbf{w}_A$ and $\mathbf{r}^{(L)} * \mathbf{w}_{A'}$ (\S\ref{sssec:method-efficient}) and the distance between their right-child templates $\mathbf{r}^{(R)} * \mathbf{w}_A$ and $\mathbf{r}^{(R)} * \mathbf{w}_{A'}$.

\paragraph{Separation of the Peak Position and the Sharpness.}
We consider the left-child distribution $\pi_{B \curvearrowleft A}$ in Equation~\eqref{eq:left_child_prob}, the probability that a parent $A$ chooses $B$ as its left child.
By Equation~\eqref{eq:corr_conv_identity}, its logit (the argument of the exponential in the numerator) is $\tau$ times the inner product of the child embedding $\mathbf{w}_B$ and the template $\mathbf{r}^{(L)} * \mathbf{w}_A$ of the parent (\S\ref{sssec:method-efficient}).
Equation~\eqref{eq:logit-decomposition} decomposes it as $\tau\, \lVert \mathbf{r}^{(L)} * \mathbf{w}_A \rVert\, \lVert \mathbf{w}_B \rVert \cos\theta_{AB}$, where $\theta_{AB}$ is the angle between $\mathbf{w}_B$ and the template.
The three factors of this expression act on the distribution as follows.
The factor $\cos\theta_{AB}$ (the closeness in direction between the child embedding and the template) determines which children receive high probability, that is, the peak position.
The factor $\tau\, \lVert \mathbf{r}^{(L)} * \mathbf{w}_A \rVert$ multiplies the logits of all children of the parent $A$ and thus determines the range of the logits, that is, the sharpness of the distribution.
The factor $\lVert \mathbf{w}_B \rVert$ multiplies the logits of the child $B$ and thus stretches or shrinks the sharpness child by child.
Hence the peak position and the sharpness are separated when (a) $\lVert \mathbf{w}_B \rVert$ is equal for all children and (b) $\lVert \mathbf{r}^{(L)} * \mathbf{w}_A \rVert$ is equal for all parents, with the embeddings determining only the peak position.

\begin{table}[t]
\centering
\scriptsize
\setlength{\tabcolsep}{3pt}
\begin{tabular}{@{}lccc@{}}
\toprule
Condition & None & Sphere & Torus \\
\midrule
(a) $\lVert \mathbf{w}_B \rVert$ equal for all children                          & $\times$ & $\checkmark$ & $\checkmark$ \\
(b) $\lVert \mathbf{r}^{(L)} * \mathbf{w}_A \rVert$ equal for all parents        & $\times$ & $\times$     & $\checkmark$ \\
\bottomrule
\end{tabular}
\caption{Whether each constraint on the vectors guarantees the two conditions under which the logit of the left-child distribution in Equation~\eqref{eq:left_child_prob} is $\cos\theta_{AB}$ multiplied by a constant ($\checkmark$) or not ($\times$). $\mathbf{w}_B$: embedding of the child $B$; $\mathbf{r}^{(L)} * \mathbf{w}_A$: template of the parent $A$. None: no constraint; sphere: all vectors, including the relation vectors, have unit $\ell_2$ norm; torus: all vectors lie on $\mathbb{T}^d$.}
\label{tab:torus-conditions}
\end{table}

Table~\ref{tab:torus-conditions} shows whether three constraints on the embeddings and the relation vectors guarantee these two conditions.
Without any constraint, as in the MLP-based scorer of SN-PCFG (\S\ref{ssec:method-torus}), neither condition holds in general.
When all vectors including the relation vectors are constrained to the unit $\ell_2$ sphere, as with the sphere projection in Table~\ref{tab:ablation-geometry}, condition (a) holds but condition (b) does not.
Circular convolution with a relation vector not on $\mathbb{T}^d$ does not preserve norms in general, so the norm of the template can vary from parent to parent.
On $\mathbb{T}^d$, every vector has unit norm and the template also lies on $\mathbb{T}^d$ by the properties above, so both conditions hold, and the logit is $\tau \cos\theta_{AB}$.
The peak position of each left-child distribution is thus determined by the template (that is, by the embeddings) and its sharpness by $\tau$.
The same holds for the right-child, lexical-emission, and root distributions in Equations~\eqref{eq:right_child_prob}, \eqref{eq:emission_prob}, and~\eqref{eq:root_prob}, whose logits $\tau \langle \mathbf{w}_C,\, \mathbf{r}^{(R)} * \mathbf{w}_A \rangle$, $\eta \langle \mathbf{u}_w,\, \mathbf{r}^{(E)} * \mathbf{w}_T \rangle$, and $\alpha \langle \mathbf{w}_A,\, \mathbf{w}_{\mathcal{S}} \rangle$ are also inner products of two vectors on $\mathbb{T}^d$ multiplied by a scaling parameter.

\section{Breakdown of Parsing Results by Label and Length}
\label{app:lang-breakdown}

\paragraph{Setup.}
We analyze the test-split trees predicted by Hol-PCFG and SN-PCFG of Tables~\ref{tab:results} and~\ref{tab:results-multilingual} under maximum-likelihood training, five random seeds each.
For the gold constituents of each label and each length, we compare the models by recall (the percentage of those constituents whose spans are in the predicted trees) averaged over the seeds.

\begin{table*}[t]
\centering
\scriptsize
\setlength{\tabcolsep}{1.8pt}
\begin{tabular}{@{}lrrrr@{\hspace{8pt}}lrrrr@{\hspace{8pt}}lrrrr@{\hspace{8pt}}lrrrr@{}}
\toprule
\multicolumn{5}{@{}c}{English} & \multicolumn{5}{c}{Chinese} & \multicolumn{5}{c}{French} & \multicolumn{5}{c@{}}{German} \\
\cmidrule(r{6pt}){1-5}\cmidrule(r{6pt}){6-10}\cmidrule(r{6pt}){11-15}\cmidrule{16-20}
Label & \# & Hol & SN & $\Delta$ & Label & \# & Hol & SN & $\Delta$ & Label & \# & Hol & SN & $\Delta$ & Label & \# & Hol & SN & $\Delta$ \\
\midrule
NP   & 14{,}346 & 84.0 & 85.0 & $-0.9$  & NP  & 1{,}865 & 66.8 & 51.3 & $+15.5$ & NP     & 13{,}726 & 67.4 & 72.1 & $-4.6$  & NP  & 10{,}673 & 81.9 & 82.9 & $-1.0$ \\
VP   & 8{,}107  & 55.8 & 50.5 & $+5.4$  & VP  & 1{,}652 & 55.8 & 42.5 & $+13.4$ & PP     & 10{,}207 & 76.2 & 73.0 & $+3.2$  & PP  & 8{,}937  & 85.7 & 85.2 & $+0.5$ \\
PP   & 5{,}482  & 82.4 & 63.3 & $+19.1$ & IP  & 637     & 59.7 & 46.4 & $+13.3$ & VN     & 2{,}879  & 46.7 & 43.7 & $+2.9$  & S   & 5{,}901  & 52.0 & 54.0 & $-2.0$ \\
S    & 3{,}539  & 68.5 & 54.8 & $+13.8$ & CP  & 400     & 57.5 & 37.9 & $+19.6$ & COORD  & 1{,}620  & 71.6 & 43.0 & $+28.6$ & VP  & 2{,}998  & 52.2 & 69.4 & $-17.2$ \\
SBAR & 1{,}792  & 69.4 & 53.8 & $+15.6$ & PP  & 337     & 72.3 & 52.3 & $+20.0$ & VPinf  & 1{,}323  & 30.6 & 32.3 & $-1.7$  & AP  & 1{,}424  & 60.6 & 37.5 & $+23.1$ \\
ADJP & 651      & 70.7 & 65.8 & $+4.9$  & DNP & 196     & 68.5 & 41.9 & $+26.5$ & Sint   & 1{,}221  & 58.8 & 52.8 & $+6.0$  & PN  & 1{,}312  & 71.9 & 76.8 & $-5.0$ \\
QP   & 489      & 77.1 & 71.5 & $+5.6$  & QP  & 120     & 65.8 & 85.0 & $-19.2$ & VPpart & 871      & 49.0 & 44.7 & $+4.3$  & CNP & 1{,}167  & 66.9 & 67.5 & $-0.6$ \\
ADVP & 286      & 75.5 & 75.0 & $+0.4$  & LCP & 112     & 69.1 & 67.1 & $+2.0$  & Ssub   & 779      & 64.4 & 54.1 & $+10.2$ & CS  & 653      & 45.3 & 42.9 & $+2.4$ \\
     &          &      &      &          &     &         &      &      &          & Srel   & 759      & 62.3 & 48.1 & $+14.2$ & AVP & 468      & 73.2 & 63.4 & $+9.8$ \\
     &          &      &      &          &     &         &      &      &          & AP     & 646      & 58.7 & 56.4 & $+2.3$  & VZ  & 416      & 56.1 & 99.4 & $-43.3$ \\
\bottomrule
\end{tabular}
\caption{Recall (\%) of gold constituents by label on the test splits, averaged over five runs. \#: number of gold constituents. $\Delta$: Hol $-$ SN\@.}
\label{tab:lang-label}
\end{table*}

\begin{table*}[t]
\centering
\scriptsize
\setlength{\tabcolsep}{3pt}
\begin{tabular}{@{}l@{\hspace{8pt}}rrrr@{\hspace{10pt}}rrrr@{\hspace{10pt}}rrrr@{\hspace{10pt}}rrrr@{}}
\toprule
 & \multicolumn{4}{c}{English} & \multicolumn{4}{c}{Chinese} & \multicolumn{4}{c}{French} & \multicolumn{4}{c@{}}{German} \\
\cmidrule(r{6pt}){2-5}\cmidrule(r{6pt}){6-9}\cmidrule(r{6pt}){10-13}\cmidrule{14-17}
Length (words) & \# & Hol & SN & $\Delta$ & \# & Hol & SN & $\Delta$ & \# & Hol & SN & $\Delta$ & \# & Hol & SN & $\Delta$ \\
\midrule
2      & 8{,}111 & 91.3 & 89.7 & $+1.6$  & 1{,}466 & 72.0 & 56.7 & $+15.3$ & 12{,}174 & 71.4 & 72.5 & $-1.1$ & 11{,}420 & 83.4 & 81.7 & $+1.7$ \\
3      & 5{,}462 & 88.0 & 83.1 & $+4.9$  & 854     & 62.9 & 50.0 & $+12.9$ & 7{,}156  & 70.0 & 71.3 & $-1.3$ & 6{,}674  & 76.0 & 76.5 & $-0.5$ \\
4      & 3{,}484 & 85.3 & 79.5 & $+5.8$  & 566     & 63.8 & 48.3 & $+15.4$ & 3{,}804  & 69.0 & 69.3 & $-0.3$ & 4{,}101  & 68.6 & 71.2 & $-2.6$ \\
5--6   & 4{,}841 & 80.7 & 73.0 & $+7.7$  & 807     & 59.3 & 45.0 & $+14.3$ & 4{,}575  & 65.9 & 65.7 & $+0.2$ & 5{,}373  & 62.4 & 65.7 & $-3.3$ \\
7--9   & 4{,}553 & 70.2 & 61.8 & $+8.4$  & 636     & 58.6 & 44.2 & $+14.4$ & 3{,}729  & 57.5 & 54.3 & $+3.1$ & 4{,}163  & 59.3 & 64.3 & $-4.9$ \\
10--14 & 4{,}211 & 55.3 & 46.5 & $+8.8$  & 503     & 57.1 & 42.1 & $+15.0$ & 3{,}190  & 45.9 & 41.5 & $+4.5$ & 2{,}350  & 57.8 & 63.1 & $-5.3$ \\
15--19 & 2{,}241 & 39.6 & 29.0 & $+10.7$ & 225     & 49.0 & 37.9 & $+11.1$ & 1{,}555  & 36.3 & 29.9 & $+6.4$ & 587      & 52.7 & 57.8 & $-5.0$ \\
20+    & 2{,}217 & 25.3 & 16.1 & $+9.2$  & 349     & 44.8 & 33.9 & $+10.8$ & 2{,}057  & 25.1 & 19.2 & $+5.9$ & 143      & 55.4 & 59.3 & $-3.9$ \\
\bottomrule
\end{tabular}
\caption{Recall (\%) of gold constituents by constituent length on the test splits, averaged over five runs.}
\label{tab:lang-length}
\end{table*}

\paragraph{Results.}
In English, Hol-PCFG and SN-PCFG recover NPs and short constituents about equally, and the recall advantage of Hol-PCFG is largest for PP (+19.1), SBAR (+15.6), and S (+13.8) and grows with constituent length (Tables~\ref{tab:lang-label} and~\ref{tab:lang-length}).
In Chinese, Hol-PCFG has higher recall for every length (+10.8 to +15.4) and for every label (+2.0 to +26.5) except QP ($-19.2$).
In French, Hol-PCFG is better on COORD (+28.6), Srel (+14.2), Ssub (+10.2), and constituents of seven or more words, and SN-PCFG on NP ($-4.6$) and slightly on constituents of 2--4 words.
Since NPs and PPs make up about 63\% and constituents of 2--4 words about 60\% of the French gold constituents, these differences largely offset each other.
In German, Hol-PCFG is better on AP (+23.1) and AVP (+9.8), whereas SN-PCFG better recovers verb phrases (VP $-17.2$, VZ $-43.3$) and constituents of four or more words.

\section{Likelihood, Sample Efficiency, and Model Capacity}
\label{app:ll-data-capacity}

\begin{table}[t]
\centering
\footnotesize
\setlength{\tabcolsep}{3pt}
\begin{tabular}{@{}lrrrrrr@{}}
\toprule
 & \multicolumn{3}{c}{Log-likelihood} & \multicolumn{3}{c@{}}{SF1} \\
\cmidrule(lr){2-4}\cmidrule(l){5-7}
Language & Hol & SN & $\Delta$ & Hol & SN & $\Delta$ \\
\midrule
English & $-108.4$ & $-105.8$ & $-2.6$ & 64.6 & 59.7 & $+4.9$ \\
Chinese & $-130.0$ & $-129.9$ & $-0.1$ & 51.8 & 40.6 & $+11.2$ \\
French  & $-131.5$ & $-130.8$ & $-0.7$ & 50.9 & 50.4 & $+0.5$ \\
German  & $-84.6$  & $-83.2$  & $-1.4$ & 49.0 & 49.4 & $-0.5$ \\
\bottomrule
\end{tabular}
\caption{Log-likelihood and SF1 of Hol-PCFG and SN-PCFG under maximum-likelihood training, averaged over five runs. $\Delta$: Hol $-$ SN.}
\label{tab:ll-sf1}
\end{table}

\begin{table}[t]
\centering
\footnotesize
\setlength{\tabcolsep}{1.5pt}
\begin{tabular}{@{}lrrrrrr@{}}
\toprule
 & \multicolumn{3}{c}{SF1} & \multicolumn{3}{c@{}}{Log-likelihood} \\
\cmidrule(lr){2-4}\cmidrule(l){5-7}
Sentences & Hol & SN & $\Delta$ & Hol & SN & $\Delta$ \\
\midrule
5k   & $49.3_{\pm 0.9}$ & $50.9_{\pm 10.3}$ & $-1.6$ & $-115.6$ & $-117.7$ & $+2.1$ \\
10k  & $55.2_{\pm 1.2}$ & $58.8_{\pm 2.1}$  & $-3.6$ & $-112.1$ & $-111.6$ & $-0.5$ \\
20k  & $61.7_{\pm 1.2}$ & $57.9_{\pm 5.7}$  & $+3.8$ & $-109.0$ & $-109.1$ & $+0.1$ \\
Full & $63.7_{\pm 0.4}$ & $58.9_{\pm 5.4}$  & $+4.9$ & $-108.5$ & $-105.6$ & $-2.9$ \\
\bottomrule
\end{tabular}
\caption{SF1 and log-likelihood on the PTB test split of Hol-PCFG and SN-PCFG trained on 5k, 10k, and 20k sentences and on the full training set, averaged over three runs. $\Delta$: Hol $-$ SN.}
\label{tab:data-size}
\end{table}

\begin{table}[t]
\centering
\footnotesize
\setlength{\tabcolsep}{3pt}
\begin{tabular}{@{}lccrrr@{}}
\toprule
Setting & $|\mathcal{N}|/|\mathcal{P}|$ & $d$ & \#params & LL & SF1 \\
\midrule
Standard                    & 4096/8192  & 512  & 11.4M & $-108.4$ & $64.6_{\pm 0.4}$ \\
$|\mathcal{N}| \times 2$    & 8192/16384 & 512  & 17.7M & $-107.6$ & $62.7_{\pm 2.4}$ \\
$d \times 2$                & 4096/8192  & 1024 & 22.8M & $-112.9$ & $63.1_{\pm 1.6}$ \\
\bottomrule
\end{tabular}
\caption{Log-likelihood (LL) and SF1 on the PTB test split of Hol-PCFG with doubled numbers of nonterminals and preterminals or a doubled embedding dimension, under maximum-likelihood training (five~runs).}
\label{tab:capacity}
\end{table}

\paragraph{Likelihood and Parsing Performance.}
The test log-likelihood is the mean per-sentence log-likelihood of the trained model on the test split.
SN-PCFG matches or exceeds Hol-PCFG in test log-likelihood on all four languages, but Hol-PCFG has higher SF1 on English and Chinese (Table~\ref{tab:ll-sf1}).
The fit to the test data thus cannot explain the SF1 differences, in line with the weak correlation between the log-likelihood and SF1 pointed out by \citet{chen2025improving} (\S\ref{ssec:bg-seminfo}).

\paragraph{Sample Efficiency.}
We compare Hol-PCFG and SN-PCFG under maximum-likelihood training on random subsets of 5k, 10k, and 20k PTB training sentences, three random seeds each.
With 5k and 10k sentences, SN-PCFG has higher SF1 than Hol-PCFG, and Hol-PCFG is ahead only with 20k sentences or more (Table~\ref{tab:data-size}).
The advantage of Hol-PCFG thus does not stem from higher sample efficiency.

\paragraph{Model Capacity.}
We evaluate Hol-PCFG under maximum-likelihood training on PTB after doubling the grammar size or the embedding dimension from the standard setting, five random seeds each.
Neither doubling raises SF1 (Table~\ref{tab:capacity}), so the SF1 of Hol-PCFG is not limited by the capacity of the HolE representation.

Together, the three comparisons suggest that the SF1 differences stem from optimization, which led Hol-PCFG and SN-PCFG to different grammars that fit the test data about equally well.

\section{Knowledge-Graph Embedding Scorers}
\label{app:kg-scorers}
This appendix details how \S\ref{ssec:disc-ablation} compares Hol-PCFG with variants whose scorers are knowledge-graph embedding methods competing with HolE.

\paragraph{Scorers.}
We replace the HolE-based score $\langle \mathbf{r},\, \mathbf{a} \star \mathbf{b} \rangle$ in Equations~\eqref{eq:left_child_prob}--\eqref{eq:emission_prob} with the scores of the other methods in Table~\ref{tab:kg-scorers}.
Everything other than the scoring is as in Hol-PCFG, including the initialization of the embeddings on the torus, their projection onto it, and the use of the scaling parameters.
Only ComplEx does not use embeddings on the torus and adds a regularizer, as described below.

\begin{table}[t]
\centering
\scriptsize
\setlength{\tabcolsep}{4pt}
\begin{tabular}{@{}llc@{}}
\toprule
\begin{tabular}[b]{@{}l@{}}Knowledge-graph\\embedding method\end{tabular} & Score & \begin{tabular}[b]{@{}c@{}}Parent--child\\symmetry\end{tabular} \\
\midrule
HolE     & $\langle \mathbf{r},\, \mathbf{a} \star \mathbf{b} \rangle$                                           & $\times$ \\
RESCAL   & $\mathbf{a}^{\top} R\, \mathbf{b}$                                                                     & $\times$ \\
ComplEx  & $\mathrm{Re}\, \langle \mathbf{r},\, \mathbf{a},\, \overline{\mathbf{b}} \rangle$                      & $\times$ \\
TransE   & $-\lVert \mathbf{a} + \mathbf{r} - \mathbf{b} \rVert$                                                  & $\checkmark$ \\
MuRE     & $-\lVert \boldsymbol{\rho} \odot \mathbf{a} + \mathbf{r} - \mathbf{b} \rVert^2 + \beta_{\mathbf{b}}$  & $\checkmark$ \\
\bottomrule
\end{tabular}
\caption{Knowledge-graph embedding methods compared in Table~\ref{tab:ablation}. $\mathbf{a}$, $\mathbf{b}$: parent and child embeddings; $\mathbf{r}$,~$R$,~$\boldsymbol{\rho}$: relation parameters of each role (left child, right child, or lexical emission); $\beta_{\mathbf{b}}$: child bias. For ComplEx, $\mathbf{a}, \mathbf{b}, \mathbf{r} \in \mathbb{C}^{d/2}$ and $\langle \mathbf{r}, \mathbf{a}, \overline{\mathbf{b}} \rangle = \sum_{k} r_k\, a_k\, \overline{b_k}$. Parent--child symmetry indicates whether the parent and the child interact symmetrically ($\checkmark$) or not ($\times$).}
\label{tab:kg-scorers}
\end{table}

\paragraph{Equivalence between HolE and ComplEx.}
HolE is equivalent to ComplEx~\citep{hayashi-shimbo-2017-equivalence}, and under this equivalence, the embeddings on the torus in Hol-PCFG correspond to complex-valued embeddings whose elements all have a modulus of 1.
Training and evaluating \mbox{ComplEx} with embeddings on the torus as for the other methods (that is, with these moduli fixed to 1) would therefore yield a model equivalent to Hol-PCFG and make the comparison meaningless.
For ComplEx, we thus follow a setting with high performance in knowledge-graph completion.
The moduli are left unconstrained, no projection onto the torus is applied, and the N3 regularizer $\lambda \sum_{\mathbf{x}} \sum_{k} |x_k|^3$~\citep{lacroix2018canonical} over the nonterminal, preterminal, word, and relation vectors is added to the training loss (maximum-likelihood or SemInfo).

\section{Learnable Scaling Parameters in SN-PCFG and SC-PCFG}
\label{app:baseline-scaling}
This appendix supplements the experiment in \S\ref{ssec:disc-ablation} in which positive learnable scaling parameters $\alpha$, $\tau$, and $\eta$ multiply the root, binary-rule, and lexical-emission scores of SN-PCFG and SC-PCFG, respectively, as in Hol-PCFG.
Each scaling parameter is initialized to 1, so training starts from the same state as the original SN-PCFG and SC-PCFG.
All other settings follow the replication runs in Table~\ref{tab:results} (public implementations and hyperparameters, with five random seeds).
As Table~\ref{tab:baseline-scaling} shows, adding the scaling parameters does not improve SF1 under either objective.

\begin{table}[t]
\centering
\scriptsize
\setlength{\tabcolsep}{4pt}
\begin{tabular}{@{}lll@{}}
\toprule
Model & LL & SemInfo \\
\midrule
SN-PCFG           & $59.7_{\pm 4.3}$  & $67.3_{\pm 0.5}$ \\
SN-PCFG $+\,\alpha, \tau, \eta$ & $58.1_{\pm 10.1}$ & $66.6_{\pm 1.1}$ \\
SC-PCFG           & $58.3_{\pm 4.4}$  & $66.2_{\pm 0.5}$ \\
SC-PCFG $+\,\alpha, \tau, \eta$ & $57.4_{\pm 3.5}$  & $64.9_{\pm 1.9}$ \\
\bottomrule
\end{tabular}
\caption{PTB test SF1 of SN-PCFG and SC-PCFG without and with learnable scaling parameters ($+\,\alpha, \tau, \eta$) over five random seeds.}
\label{tab:baseline-scaling}
\end{table}

\section{Comparison with Reduced SN-PCFG Rule Scorers}
\label{app:reduced-sn}
Supplementing \S\ref{ssec:disc-params}, we shrink the SN-PCFG rule scorer with all other components unchanged and compare SF1 at each parameter count.

\paragraph{Reduced Scorers.}
SN-PCFG~\citep{liu-etal-2023-simple} scores binary and lexical rules from symbol embeddings transformed by four residual MLPs ${\mathbf{x} \mapsto \mathbf{x} + \mathrm{MLP}(\mathbf{x})}$, one each for the parent, left child, right child, and preterminal, totaling 2{,}626{,}560 parameters.
We keep the residual connections and replace each MLP with one of four maps $g$:
(i) one layer, $g(\mathbf{x}) = \mathrm{ReLU}(W\mathbf{x} + \mathbf{b})$ with $W \in \mathbb{R}^{d \times d}$, which retains the one-layer parent and child MLPs and reduces the seven-layer preterminal MLP to one layer;
(ii) rank 16, ${g(\mathbf{x}) = \mathrm{ReLU}(UV^{\top}\mathbf{x})}$ with $U, V \in \mathbb{R}^{d \times 16}$;
(iii) rank 1, $g(\mathbf{x}) = \mathrm{ReLU}(\mathbf{u}\mathbf{v}^{\top}\mathbf{x})$ with $\mathbf{u}, \mathbf{v} \in \mathbb{R}^{d}$; and
(iv) diagonal, $g(\mathbf{x}) = \mathrm{ReLU}(\mathbf{s} \odot \mathbf{x})$ with $\mathbf{s} \in \mathbb{R}^{d}$.
Variants (ii)--(iv) have no bias, and variants (iii) and (iv) have parameter counts ($\#\pi$) of the same order as Hol-PCFG (Table~\ref{tab:reduced-sn}).

\begin{table}[t]
\centering
\scriptsize
\setlength{\tabcolsep}{3pt}
\begin{tabular}{@{}llrll@{}}
\toprule
Rule scorer & $g(\mathbf{x})$ & $\#\pi$ & LL & SemInfo \\
\midrule
SN-PCFG       & (original)                                            & 2{,}626{,}560 & $59.7_{\pm 4.3}$ & $67.3_{\pm 0.5}$ \\
\midrule
(i) One layer & $\mathrm{ReLU}(W\mathbf{x} + \mathbf{b})$             & 1{,}050{,}624 & $53.0_{\pm 2.2}$ & $66.7_{\pm 0.8}$ \\
(ii) Rank 16  & $\mathrm{ReLU}(UV^{\top}\mathbf{x})$                  & 65{,}536      & $55.0_{\pm 8.0}$ & $66.3_{\pm 0.8}$ \\
(iii) Rank 1  & $\mathrm{ReLU}(\mathbf{u}\mathbf{v}^{\top}\mathbf{x})$ & 4{,}096       & $46.1_{\pm 4.5}$ & $65.2_{\pm 1.1}$ \\
(iv) Diagonal & $\mathrm{ReLU}(\mathbf{s} \odot \mathbf{x})$          & 2{,}048       & $50.6_{\pm 1.9}$ & $61.7_{\pm 2.9}$ \\
\midrule
Hol-PCFG      & ---                                                   & 1{,}539       & $64.6_{\pm 0.4}$ & $68.1_{\pm 0.5}$ \\
\bottomrule
\end{tabular}
\caption{PTB test SF1 of SN-PCFG with reduced rule scorers (five random seeds).}
\label{tab:reduced-sn}
\end{table}

\paragraph{Results.}
Every reduced variant falls below the full SN-PCFG under both objectives (Table~\ref{tab:reduced-sn}).
Variants (iii) and (iv) underperform Hol-PCFG by 14.0--18.5 points under maximum-likelihood training and by 2.9--6.4 points under SemInfo, versus 4.9 and 0.8 points for the full SN-PCFG.
Shrinking the MLP-based scorer of SN-PCFG thus does not yield the parameter efficiency of the HolE-based scorer.

\section{Joint Scoring of the Two Children}
\label{app:joint-scoring}
This appendix supplements the discussion of the conditional independence assumption in \S\ref{ssec:disc-params}.

\paragraph{Combinatorial Explosion of the Rule Table.}
As noted in \S\ref{ssec:bg-npcfg}, without the conditional independence of the two children in Equation~\eqref{eq:factorized_rule}, $\pi_{A \to BC}$ must be held for every triple $(A, B, C)$, that is, $|\mathcal{N}|(|\mathcal{N}|+|\mathcal{P}|)^2$ values.
The assumption factorizes $\pi_{A \to BC}$ into left- and right-child terms that together need only $2|\mathcal{N}|(|\mathcal{N}|+|\mathcal{P}|)$ values, a reduction by a factor of $(|\mathcal{N}|+|\mathcal{P}|)/2$.
A parameter-efficient scorer does not change this ratio, because it reduces the parameters that produce the table, not the table itself.
Modeling the dependence between the two children therefore requires a small grammar, on the order of $|\mathcal{N}| = 30$ and $|\mathcal{P}| = 60$.

\paragraph{Effect of Modeling the Dependence.}
We evaluate the effect of modeling the dependence between the two children in a small grammar, using three Hol-PCFG variants that differ only in how $\pi_{A \to BC}$ is computed:
\begin{itemize}
  \item \emph{Factorized}: the unmodified Hol-PCFG (Equations~\eqref{eq:factorized_rule}, \eqref{eq:left_child_prob}, and~\eqref{eq:right_child_prob}).
  \item \emph{Factorized\,+\,interaction}: adds an interaction term between the two children (a HolE score that composes the parent and both children) to the scores of the factorized variant and normalizes over all child pairs $(B, C)$:
  \begin{equation}
  \begin{aligned}
    &\pi_{A \to BC} \propto \exp\bigl(\tau \langle \mathbf{r}^{(L)}, \mathbf{w}_A \star \mathbf{w}_B \rangle \\
    &\quad {}+ \tau \langle \mathbf{r}^{(R)}, \mathbf{w}_A \star \mathbf{w}_C \rangle \\
    &\quad {}+ \kappa \langle \mathbf{v}, \mathbf{w}_A \star (\mathbf{w}_B \star \mathbf{w}_C) \rangle\bigr),
  \end{aligned}
  \label{eq:interaction-fact}
  \end{equation}
  with a relation vector $\mathbf{v} \in \mathbb{T}^d$ for the interaction term and a scaling parameter $\kappa > 0$.
  \item \emph{Interaction only}: the interaction term alone:
  \begin{equation}
    \pi_{A \to BC} \propto \exp\bigl(\tau \langle \mathbf{v}, \mathbf{w}_A \star (\mathbf{w}_B \star \mathbf{w}_C) \rangle\bigr).
    \label{eq:interaction-only}
  \end{equation}
\end{itemize}
With the interaction term, the two children are in general no longer conditionally independent given the parent.
Under maximum-likelihood training on PTB, we compare all variants with $|\mathcal{N}| = 30$ and $|\mathcal{P}| = 60$ by SF1 over five random seeds.

The variants with the interaction term improve SF1 by about 1.6--2.6 points over the factorized variant, whereas scaling the factorized variant up to $|\mathcal{N}| = 4096$ and $|\mathcal{P}| = 8192$ adds 22.5 points (Table~\ref{tab:joint-scoring}).
For HolE-based scoring, capturing the dependence between the children thus has a far smaller effect than the grammar size that the factorization makes possible.

\begin{table}[t]
\centering
\small
\setlength{\tabcolsep}{4pt}
\begin{tabular}{@{}lcrc@{}}
\toprule
Binary-rule scorer & $|\mathcal{N}|/|\mathcal{P}|$ & $\#\pi$ & SF1 \\
\midrule
Factorized             & 30/60     & 1{,}025 & $42.1_{\pm 1.6}$ \\
Factorized + interaction & 30/60   & 1{,}538 & $44.7_{\pm 2.1}$ \\
Interaction only         & 30/60   & 513     & $43.7_{\pm 5.7}$ \\
\midrule
Factorized (Hol-PCFG)  & 4096/8192 & 1{,}025 & $64.6_{\pm 0.4}$ \\
\bottomrule
\end{tabular}
\caption{PTB test SF1 of Hol-PCFG variants with and without the conditional independence of the two children, under maximum-likelihood training (five random seeds). The last row is Hol-PCFG in Table~\ref{tab:results}. $\#\pi$: parameters of the binary-rule scorer.}
\label{tab:joint-scoring}
\end{table}

\paragraph{Non-Associativity in the Interaction Term.}
Circular correlation is in general non-associative ($(\mathbf{a} \star \mathbf{b}) \star \mathbf{c} \neq \mathbf{a} \star (\mathbf{b} \star \mathbf{c})$).
The interaction terms of Equations~\eqref{eq:interaction-fact} and~\eqref{eq:interaction-only} compose three vectors, so the representable rule distributions may change with the order of the operations.
For $\mathbf{x} \in \mathbb{T}^d$, we write $\hat{x}_k = e^{i\phi^{x}_k}$ (\S\ref{ssec:method-torus}) and denote the phases of $\mathbf{w}_X$ by $\phi^{X}_k$.
Since $\mathcal{F}[\mathbf{a} \star \mathbf{b}]_k = \overline{\hat{a}_k}\, \hat{b}_k$, if $\mathbf{a}, \mathbf{b} \in \mathbb{T}^d$, then $|\mathcal{F}[\mathbf{a} \star \mathbf{b}]_k| = 1$ for every $k$.
Hence $\mathbf{a} \star \mathbf{b} \in \mathbb{T}^d$, with phases $\phi^{b}_k - \phi^{a}_k$.
The vectors composed in the two orders are thus both on $\mathbb{T}^d$, with the phases
\begin{align}
  \mathbf{w}_A \star (\mathbf{w}_B \star \mathbf{w}_C) &\;\mapsto\; \phi^{C}_k - \phi^{B}_k - \phi^{A}_k, \label{eq:phase-orig} \\
  (\mathbf{w}_A \star \mathbf{w}_B) \star \mathbf{w}_C &\;\mapsto\; \phi^{C}_k - \phi^{B}_k + \phi^{A}_k, \label{eq:phase-other}
\end{align}
which differ only in the sign of $\phi^{A}_k$.
For any $\mathbf{x} \in \mathbb{T}^d$, Parseval's theorem (for real vectors and the unnormalized DFT, $\langle \mathbf{a}, \mathbf{b} \rangle = \frac{1}{d} \sum_k \overline{\hat{a}_k}\, \hat{b}_k$) gives $\langle \mathbf{v}, \mathbf{x} \rangle = \frac{1}{d} \sum_k \overline{\hat{v}_k}\, \hat{x}_k$, that is,
\begin{equation*}
  \langle \mathbf{v}, \mathbf{x} \rangle
  = \frac{1}{d} \sum_k e^{-i\phi^{v}_k} e^{i\phi^{x}_k}
  = \frac{1}{d} \sum_k e^{i(\phi^{x}_k - \phi^{v}_k)}.
\end{equation*}
The left-hand side is real, being the inner product of two real vectors.
Since the real part of $e^{i\psi}$ is $\cos\psi$ (Euler's formula), taking real parts gives
\begin{equation}
  \langle \mathbf{v}, \mathbf{x} \rangle = \frac{1}{d} \sum_k \cos(\phi^{x}_k - \phi^{v}_k).
  \label{eq:torus-score}
\end{equation}
Let $s(A, B, C; \mathbf{v}) = \langle \mathbf{v}, \mathbf{w}_A \star (\mathbf{w}_B \star \mathbf{w}_C) \rangle$ be the score of the interaction term in the original order and $s'(A, B, C; \mathbf{v}) = \langle \mathbf{v}, (\mathbf{w}_A \star \mathbf{w}_B) \star \mathbf{w}_C \rangle$ that in the other order.
Substituting the phases in Equations~\eqref{eq:phase-orig} and~\eqref{eq:phase-other} for $\phi^{x}_k$ in Equation~\eqref{eq:torus-score} gives
\begin{align}
  &s(A, B, C; \mathbf{v}) \notag \\
  &\;= \frac{1}{d} \sum_k \cos(\phi^{C}_k - \phi^{B}_k - \phi^{A}_k - \phi^{v}_k), \label{eq:score-orig} \\
  &s'(A, B, C; \mathbf{v}) \notag \\
  &\;= \frac{1}{d} \sum_k \cos(\phi^{C}_k - \phi^{B}_k + \phi^{A}_k - \phi^{v}_k). \label{eq:score-other}
\end{align}
Since cosine is an even function, negating the entire argument of the cosine in Equation~\eqref{eq:score-other} gives
\begin{align}
  &s'(A, B, C; \mathbf{v}) \notag \\
  &\;= \frac{1}{d} \sum_k \cos\bigl((\phi^{B}_k - \phi^{C}_k - \phi^{A}_k) - (-\phi^{v}_k)\bigr). \label{eq:score-other-flip}
\end{align}
In Equation~\eqref{eq:score-other-flip}, $\phi^{B}_k - \phi^{C}_k - \phi^{A}_k$ is the phase in Equation~\eqref{eq:phase-orig} with $B$ and $C$ swapped, and $-\phi^{v}_k$ is the phase of $\tilde{\mathbf{v}}$, defined by negating the phases of $\mathbf{v}$.
Negation preserves unit modulus and Hermitian symmetry, so $\tilde{\mathbf{v}}$ is a real vector on $\mathbb{T}^d$.
The right-hand side of Equation~\eqref{eq:score-other-flip} is that of Equation~\eqref{eq:score-orig} with $B$ and $C$ swapped and $\mathbf{v}$ replaced by $\tilde{\mathbf{v}}$:
\begin{equation}
  s'(A, B, C; \mathbf{v}) = s(A, C, B; \tilde{\mathbf{v}}).
  \label{eq:order-mirror}
\end{equation}
By Equation~\eqref{eq:order-mirror}, the score that the model with the other order assigns to $(A, B, C)$ under the relation vector $\mathbf{v}$ equals the score that the original model assigns to $(A, C, B)$ under $\tilde{\mathbf{v}}$.
In the factorized\,+\,interaction variant, the two factorized terms of Equation~\eqref{eq:interaction-fact} for $(A, B, C)$ likewise equal those for $(A, C, B)$ with $\mathbf{r}^{(L)}$ and $\mathbf{r}^{(R)}$ exchanged.
Since the relation vectors are parameters learned on $\mathbb{T}^d$ and $\tilde{\mathbf{v}}$ also lies on $\mathbb{T}^d$, the original model can take $\tilde{\mathbf{v}}$ as its relation vector, with $\mathbf{r}^{(L)}$ and $\mathbf{r}^{(R)}$ exchanged.
Hence every rule distribution represented by the model with the other order is a rule distribution of the original model with the two children swapped.
Thus, changing the order only mirrors the representable rule distributions left to right.

\section{Details of the Analyses of Induced Nonterminals}
\label{app:nt-hierarchy}
This appendix describes how the nearest-neighbor agreement in Table~\ref{tab:nn-agreement} and the dendrogram in Figure~\ref{fig:nt-hierarchy} (\S\ref{ssec:disc-hierarchy}) are computed.

\paragraph{Nearest-Neighbor Agreement in Table~\ref{tab:nn-agreement}.}
We analyze Hol-PCFG and SN-PCFG trained on PTB with the maximum-likelihood objective (five seeds each).
On the PTB dev split, differentiating the log-partition function of each model computed with the inside algorithm gives the posterior probability that each span is a constituent labeled with each nonterminal.
Summing these probabilities over the split gives the usage $u(A)$ of a nonterminal~$A$, the expected number of constituents it generates.
We analyze the nonterminals with $u(A) \geq 1$ (on average, about 2{,}100 for Hol-PCFG and 1{,}300 for SN-PCFG).

The rule-distribution distance between two nonterminals is the Euclidean distance between the vectors concatenating the square roots of the left- and right-child probabilities of each nonterminal (a Hellinger-type distance).
The embedding distance for Hol-PCFG is the Euclidean distance between the embeddings $\mathbf{w}_A$, equal to the distance between their templates (\S\ref{ssec:method-torus}).
SN-PCFG computes each child logit as the inner product of a parent-side vector $f(\mathbf{w}_A)$ and a child-side vector, where $f$ is a residual MLP~\citep{liu-etal-2023-simple}.
For SN-PCFG, we therefore use three distances, namely the Euclidean distances between the embeddings $\mathbf{w}_A$ and between the parent-side vectors, and the cosine distance between the parent-side vectors, which ignores their norms.
The agreement is the fraction of the ten nearest neighbors of a nonterminal under the embedding distance that are among its ten nearest neighbors under the rule-distribution distance, averaged over the \mbox{nonterminals}.

\paragraph{Dendrogram in Figure~\ref{fig:nt-hierarchy}.}
We analyze Hol-PCFG trained on PTB with the SemInfo objective ($|\mathcal{N}| = 1024$, $|\mathcal{P}| = 2048$, $d = 512$).
By the same computation, 892 of its 1{,}024 nonterminals have $u(A) \geq 1$ and account for 99.8\% of the total usage.
Since the embedding distance equals the template distance, we apply Ward agglomerative clustering directly to the embeddings $\mathbf{w}_A$ of these nonterminals, with optimal leaf ordering.
To relate the dendrogram to treebank categories, we sum the posterior probability of nonterminal $A$ over the spans of the gold constituents of category~$L$ (width $\geq 2$).
This label mass $\nu(L, A)$ distributes the gold constituents of $L$ over the nonterminals by the model posterior.
Each link is colored by the category with the largest share of the label mass below it, or gray if that share is below 50\%.
For each category $L$, the shaded box, the dot, and the bar mark the subtree rooted at the node $v$ with the highest F1, the harmonic mean of the share of $L$ in the label mass below $v$ (precision) and the fraction of the total label mass of $L$ that lies below $v$~(recall).

\end{document}